\documentclass[iicol,pdflatex,sn-mathphys-ay]{sn-jnl}% Chicago-based Humanities Reference Style

\usepackage{graphicx}%
\usepackage{multirow}%
\usepackage{amsmath,amssymb,amsfonts}%
\usepackage{amsthm}%
\usepackage{mathrsfs}%
\usepackage[title]{appendix}%
\usepackage{xcolor}%
\usepackage{textcomp}%
\usepackage{manyfoot}%
\usepackage{booktabs}%
\usepackage{algorithm}%
\usepackage{algorithmicx}%
\usepackage{algpseudocode}%
\usepackage{listings}%

\usepackage{booktabs}
\usepackage{rotating} 
\usepackage{pifont}
\usepackage{multirow}
\usepackage{multicol}
\usepackage{makecell}
\usepackage{colortbl}
\definecolor{ours}{RGB}{225, 235, 246}
\usepackage{wasysym}
\usepackage{utfsym}
\usepackage{fontawesome}
\usepackage{afterpage}
\usepackage{placeins}
\usepackage{enumitem}
\usepackage{lipsum} % for filler text
\usepackage{hhline}
\theoremstyle{thmstyleone}%
\theoremstyle{thmstyletwo}%

\theoremstyle{thmstylethree}%

\begin{document}

\title[Article Title]{Dinomaly2: A Unified Framework for Unsupervised Image Anomaly Detection}


% \author*[1,2]{\fnm{First} \sur{Author}}\email{iauthor@gmail.com}



% \affil*[1]{\orgdiv{Department}, \orgname{Organization}, \orgaddress{\street{Street}, \city{City}, \postcode{100190}, \state{State}, \country{Country}}}

% \affil[2]{\orgdiv{Department}, \orgname{Organization}, \orgaddress{\street{Street}, \city{City}, \postcode{10587}, \state{State}, \country{Country}}}

% \affil[3]{\orgdiv{Department}, \orgname{Organization}, \orgaddress{\street{Street}, \city{City}, \postcode{610101}, \state{State}, \country{Country}}}

%% ================= Athors =================
% 1: Tsinghua University
\author[1]{\fnm{Jia} \sur{Guo}}

% 2: Beijing Institute of Technology
\author[2]{\fnm{Shuai} \sur{Lu}}

% 3: University of New South Wales
\author[3]{\fnm{Lei} \sur{Fan}}

% 4: City University of Hong Kong
\author[4]{\fnm{Zelin} \sur{Li}}

% 1: Tsinghua, 5: DZ Matrix
\author[1,5]{\fnm{Donglin} \sur{Di}}

% 3: University of New South Wales
\author[3]{\fnm{Yang} \sur{Song}}

% 2: Beijing Institute of Technology
\author[2]{\fnm{Weihang} \sur{Zhang}}

% 6: Fudan University, 7: Rongcheer Co., Ltd.
\author[6,7]{\fnm{Wenbing} \sur{Zhu}}

% 4: City University of Hong Kong
\author[4]{\fnm{Hong} \sur{Yan}}

% 8: Shanghai Jiao Tong University (Corresponding Author)
\author*[8]{\fnm{Fang} \sur{Chen}}\email{chen-fang@sjtu.edu.cn}

% 2: Beijing Institute of Technology (Corresponding Author)
\author*[2]{\fnm{Huiqi} \sur{Li}}\email{huiqili@bit.edu.cn}

% 1: Tsinghua, 8: SJTU (Corresponding Author)
\author*[1,8]{\fnm{Hongen} \sur{Liao}}\email{liao@tsinghua.edu.cn}

%% ================= Schools =================
\affil[1]{\orgname{Tsinghua University}, \orgaddress{\city{Beijing}, \country{China}}}

\affil[2]{\orgname{Beijing Institute of Technology}, \orgaddress{\city{Beijing}, \country{China}}}

\affil[3]{\orgname{University of New South Wales}, \orgaddress{\city{Sydney}, \country{Australia}}}

\affil[4]{\orgname{City University of Hong Kong}, \orgaddress{\city{Hong Kong SAR}, \country{China}}}

\affil[5]{\orgname{DZ Matrix}, \orgaddress{\country{China}}}

\affil[6]{\orgname{Fudan University}, \orgaddress{\city{Shanghai}, \country{China}}}

\affil[7]{\orgname{Rongcheer Co., Ltd.}, \orgaddress{\country{China}}}

\affil[8]{\orgname{Shanghai Jiao Tong University}, \orgaddress{\city{Shanghai}, \country{China}}}


\abstract{
 Unsupervised anomaly detection (UAD) has evolved from specialized single-class models to unified multi-class models, yet existing multi-class models significantly underperform the most advanced one-for-one counterparts. Moreover, the field has fragmented into specialized methods tailored to specific scenarios (multi-class, 3D, few-shot, etc.), creating deployment barriers and highlighting the need for a unified solution. In this paper, we present Dinomaly2, a unified framework for image UAD. It bridges the performance gap in multi-class models by addressing over-generalization and contextual ambiguity, while extending across diverse data modalities and task settings.
Guided by the ``less is more'' philosophy, we demonstrate that the orchestration of simple elements—foundational representations, basic dropouts, simplified attention mechanisms, contextual recentering, and loosened optimization objectives—achieves superior performance in a standard reconstruction-based framework. This methodological minimalism enables natural extension across diverse tasks with minimal task-specific adaptations, establishing that simplicity is the foundation of unification. Extensive experiments on 15 UAD benchmarks demonstrate strong performance across multiple {modalities}, {task settings}, and {application domains}.
For example, our unified multi-class model achieves unprecedented 99.9\% and 99.3\% image-level (I-) AUROC on MVTec-AD and VisA, respectively, with its effectiveness further extending to large-scale datasets containing hundreds of categories. Moreover, using only 8 normal examples per class, our method surpasses previous full-shot models, achieving 98.7\% and 97.4\%. For multi-view (Real-IAD, MANTA) and multi-modal (MVTec3D, MulSen-AD) inspection, Dinomaly2 demonstrates state-of-the-art performance through lightweight adaptations.  The combination of minimalistic design, computational scalability, and broad applicability positions Dinomaly2 as a shared framework for a wide range of real-world anomaly detection applications.
}




\maketitle

\section{Introduction}\label{sec1}

\begin{figure*}[!t]
\centerline{\includegraphics[width=0.96\textwidth]{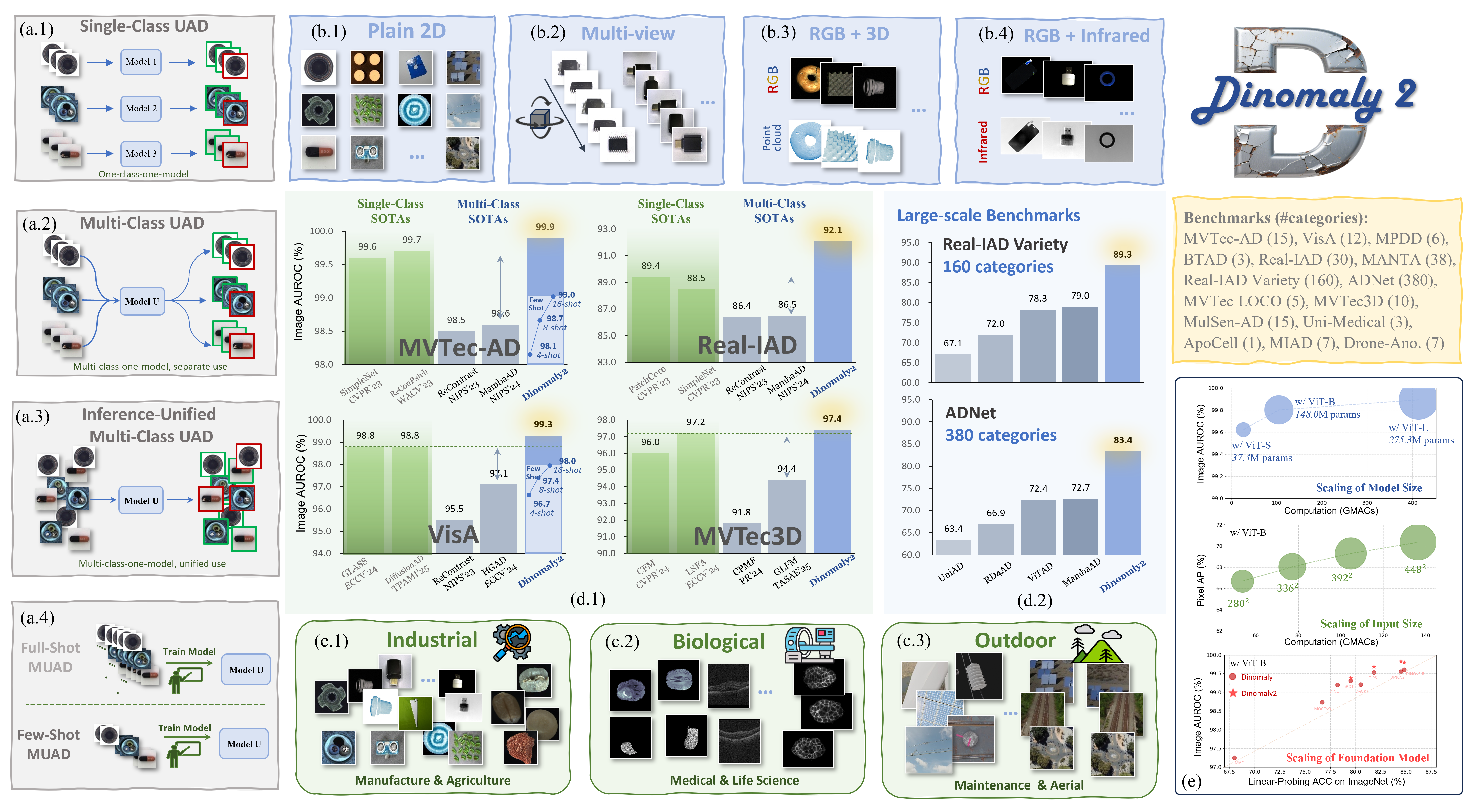}}
\caption{Overview: task settings, benchmarks, and scalability of Dinomaly2. (a) Task settings include single-class UAD, multi-class UAD (MUAD), inference-unified MUAD, and few-shot MUAD. (b) Multi-modal capabilities span 2D images, multi-view 2D, RGB+3D, and RGB+Infrared. (c) Cross-domain validation across industrial, biological, and outdoor. (d)  Dinomaly2 achieves state-of-the-art performance on MVTec-AD (15 categories), VisA (12 categories), Real-IAD (30 categories$\times$5 views), and MVTec3D (10 categories). This effectiveness further extends to large-scale datasets, including Real-IAD Variety (160 categories) and ADNet (380 categories). (e) Scalability analysis shows consistent improvements with model size, input resolution, and foundation model quality (on MVTec-AD).}
\label{fig1}
\end{figure*}

Sensory unsupervised anomaly detection (UAD) aims to identify anomalous regions with covariate shifts in images without prior knowledge of anomaly types \citep{yang2024generalized}, serving as a fundamental task in computer vision with applications ranging from industrial inspection ~\citep{bergmann2019mvtec,wang2024real,bergmann2022beyond}, medical diagnosis ~\citep{xiang2024exploiting,guo2023encoder}, to surveillance systems ~\citep{mabrouk2018abnormal}. The inherent scarcity and diversity of anomalies make this task particularly challenging, as models must learn to represent normal patterns from limited data while remaining sensitive to deviations during inference.

Early image UAD approaches follow a one-class-one-model paradigm~\citep{roth2022towards,deng2022anomaly}, where separate models are trained for each object category or scene (Fig.~\ref{fig1}(a.1)). While this strategy has proven effective in controlled scenarios, it faces significant challenges in real-world applications. The storage overhead, computational complexity, and maintenance burden of managing hundreds of individual models become prohibitive in practical deployments. Recent advances have shifted toward \textit{multi-class UAD} (MUAD, Fig.~\ref{fig1}(a.2)), where a single unified model is designed to handle multiple object categories simultaneously~\citep{you2022unified,zhao2023omnial,guo2023recontrast}.
Recent methods leverage a variety of techniques, including neighbor-masked attention~\citep{you2022unified}, synthetic anomalies~\citep{zhao2023omnial}, vector quantization~\citep{lu2023hierarchical}, diffusion models~\citep{yin2023lafite,he2024diad}, and state space models such as Mamba~\citep{he2024mambaad} to capture compact representations of normal patterns, which are critical for distinguishing anomalies. However, as normal patterns become increasingly diverse with the inclusion of more categories, the underlying distribution becomes difficult to characterize, consequently degrading detection performance~\citep{you2022unified}.
Yet despite this proliferation of complex designs, there is still a non-negligible performance gap between the state-of-the-art (SOTA) multi-class and single-class models (Fig.~\ref{fig1}(d)), which restricts the practical adoption of unified approaches.

Furthermore, existing UAD approaches suffer from a more fundamental problem: \textit{methodological fragmentation}. The field has evolved into a collection of specialized frameworks, each tailored to specific task settings. For example, methods designed for plain 2D inspection~\citep{you2022unified,zhao2023omnial} cannot be naturally extended to multi-modal data without modifications ~\citep{horwitz2023back,wang2023multimodal}. Methods designed for few-shot UAD often require entirely new pipelines that leverage vision-language models or meta-learning ~\citep{jeong2023winclip,zhu2024toward}. This proliferation of specialized methods creates huge barriers for practitioners. They must master multiple frameworks, maintain disparate codebases, and manage incompatible architectures when deploying anomaly detection across different scenarios. As UAD applications continue to diversify, the demand for unified frameworks becomes paramount. 

In this work, we aim to challenge the prevailing trend toward architectural complexity with a counterintuitive proposition: unified anomaly detection requires simplification, not specialization.
Here we present \textbf{Dinomaly2}, a unified framework for multi-class, multi-modal, and multi-task UAD, built on principled minimalism. 

\begin{figure*}[!t]
\centerline{\includegraphics[width=\textwidth]{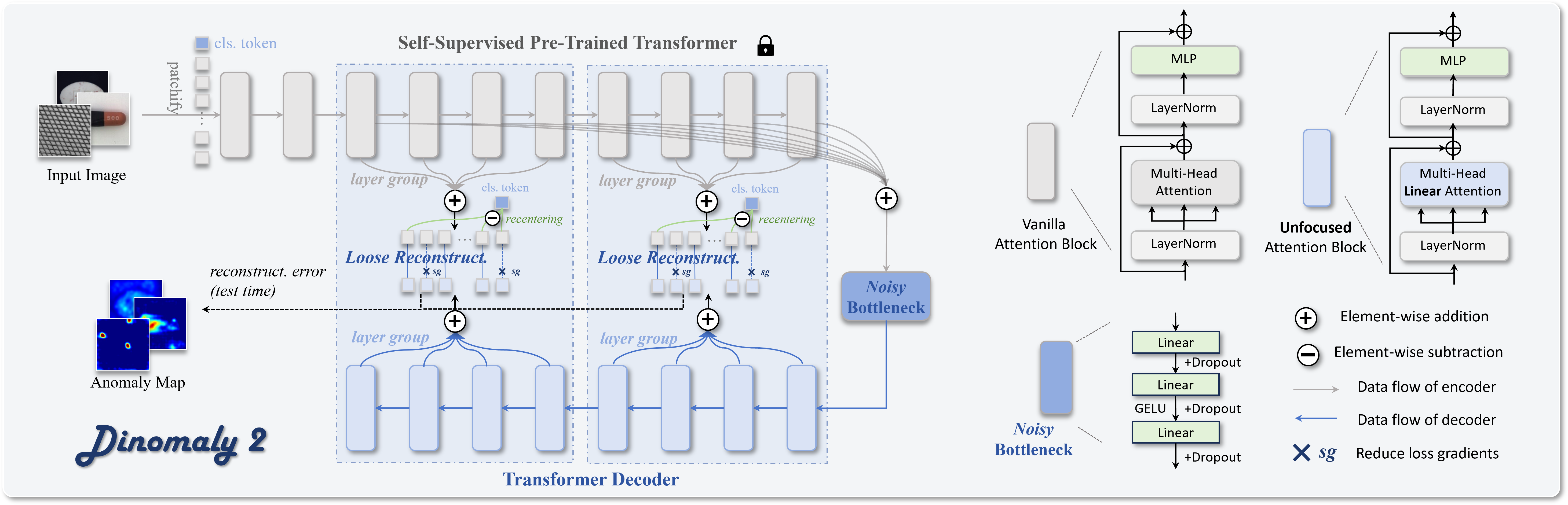}}
\caption{The framework of Dinomaly2, built by simple and pure Transformer building blocks. A pre-trained encoder extracts multi-layer features, which are aggregated through a bottleneck and reconstructed by a decoder. During training, the model is optimized exclusively on normal images, while during inference, anomalies are detected based on reconstruction errors. }
\label{fig2}
\end{figure*}

Based on the epistemic behavior of neural networks, Dinomaly2 detects anomalous regions via reconstruction error (Fig.~\ref{fig2}). We systematically analyze two challenges in extending this paradigm to unified multi-class UAD: \textit{over-generalization}, where reconstruction ability learned from normal data extends to anomalous patterns, and \textit{contextual ambiguity}, where the normality of a local pattern depends on its surrounding context. To address these challenges, Dinomaly2 follows the \textbf{\textit{``less is more''}} philosophy and comprises five simple yet essential elements:

\begin{itemize}[left=0pt]
\item \textit{Foundational representations:} We demonstrate that self-supervised Vision Transformers (ViTs), when properly scaled, provide transferable feature representations that generalize across diverse UAD modalities and domains.
\item \textit{Less information:} We propose a \textit{Noisy Bottleneck} that activates the built-in Dropout in an MLP to prevent over-generalization, as an alternative to carefully designed synthetic anomalies and feature perturbations.
\item \textit{Less attention:} We exploit \textit{Unfocused Linear Attention}'s inherent inability to focus as a feature rather than a limitation, preventing the propagation of identical information during reconstruction.
\item \textit{Less constraint:} We propose \textit{Loose Reconstruction}, which deliberately relaxes layer-to-layer and point-by-point correspondence to prevent the decoder from perfectly mimicking the encoder.
\item  \textit{Less context invariance:} We introduce \textit{Context-Aware Recentering}, which subtracts CLS tokens from patch features to resolve multi-class confusion, where the same feature pattern may be normal in one context but anomalous in another.
\end{itemize}

This methodological minimalism enables Dinomaly2 to extend naturally
	across diverse modalities and task settings, while preserving Dinomaly2's core reconstruction architecture and mechanism. Specifically, we introduce
	lightweight task-specific adaptations for multi-view inspection, RGB+3D point clouds,
	RGB+Infrared (IR) images, and few-shot UAD (Fig.~\ref{fig1}(b.2-4, a.4)). In addition, we demonstrate Dinomaly2's robustness by introducing the \textit{inference-unified MUAD} setting (Fig.~\ref{fig1}(a.3)), where anomalies are detected across mixed categories for truly unified deployment. 

We validate Dinomaly2's broad applicability on 15 benchmark datasets covering 692 categories/scenes (Table~\ref{tab:anomaly_datasets}), spanning four data modalities (2D, multi-view 2D, RGB-3D, RGB-IR), four task settings (single-class UAD, MUAD, inference-unified MUAD, few-shot), and three application domains (industrial, biological, outdoor).
Dinomaly2 achieves strong unified detection performance across industrial inspection settings and category scales. On conventional industrial benchmarks, it achieves 99.9\%, 99.3\%, and 92.1\% image-level (I-) AUROC on MVTec-AD~\citep{bergmann2019mvtec}, VisA~\citep{zou2022spot}, and Real-IAD~\citep{wang2024real}, respectively, surpassing both multi-class and single-class SOTAs (Fig.~\ref{fig1}(d)). For multi-view inspection, it achieves 94.9\% and 94.6\% object-level AUROC on Real-IAD and MANTA~\citep{fan2025manta}. This effectiveness extends to hundreds of categories, with 89.3\% and 83.4\% I-AUROC on Real-IAD Variety~\citep{zhu2026real} (160 categories) and ADNet~\citep{ling2026adnet} (380 categories), respectively. With lightweight multi-modal adaptations, Dinomaly2 also outperforms specialized methods, achieving 97.4\% I-AUROC on MVTec3D~\citep{bergmann2021mvtec3d} and 97.6\% object-level AUROC on MulSen-AD~\citep{li2025multi}.

Beyond conventional industrial inspection, Dinomaly2 generalizes across task settings and application domains. Under few-shot training, Dinomaly2 achieves 98.7\% and 97.4\% I-AUROC on MVTec-AD and VisA, respectively, using only eight normal examples per class. Under inference-unified evaluation, it retains strong performance when categories are mixed without class-specific thresholds, achieving 98.9\% and 97.8\% I-AUROC, respectively. In addition, we evaluate it on medical images~\citep{bao2024bmad}, microscopic images for life science~\citep{guan2025cell}, outdoor maintenance~\citep{bao2023miad}, and surveillance drone imagery~\citep{jin2022anomaly}. We further systematically investigate its scalability, showing positive scaling behavior with model size, input resolution, and foundation-model quality (Fig.~\ref{fig1}(e)).

A preliminary version of this work was published at CVPR 2025 (Dinomaly~\citep{guo2025dinomaly}). This work extends the initial version into a more comprehensive and powerful framework with the following major aspects: (1) Context-Aware Recentering, which resolves multi-class confusion through feature space transformation; (2) Natural extension to various modalities and settings with lightweight adaptations, including multi-view, RGB-3D, RGB-IR, and few-shot UAD; (3) Introduction and validation of the inference-unified MUAD setting, advancing toward complete unification where models operate across mixed categories; and (4) Comprehensive validation across 15 benchmark datasets spanning various modalities, settings, and domains, establishing Dinomaly2 as a unified anomaly detection framework. 

\section{Related Work}

\textbf{Image Unsupervised Anomaly Detection} has evolved through several complementary paradigms.
\textit{Feature reconstruction} (or distillation) methods~\citep{deng2022anomaly,salehi2021multiresolution,hyun2024reconpatch} train a student decoder to reconstruct features extracted by a pre-trained encoder, detecting anomalies where reconstruction fails.
\textit{Synthetic anomaly} methods~\citep{li2021cutpaste,zavrtanik2021draem,zhang2023destseg,liu2023simplenet} generate artificial defects on normal images and convert UAD into a supervised segmentation task, but their effectiveness depends heavily on how well the synthetic anomalies approximate real ones.
\textit{Memory-based} methods~\citep{defard2021padim,roth2022towards,lee2022cfa} store normal features from pre-trained networks and detect anomalies via distribution matching at inference. Such methods incur substantial memory and computation costs that scale with training set size.
Recent works have also explored normalizing flows~\citep{gudovskiy2022cflow,yu2021fastflow} and diffusion models~\citep{he2024diad,zhang2025diffusionad} for modeling normal distributions.

\textbf{Multi-Class UAD.}
UniAD~\citep{you2022unified} first formalized the multi-class setting and proposed to prevent identity mapping (over-generalization).
Subsequent works tackled over-generalization from various angles: OmniAL~\citep{zhao2023omnial} via synthetic anomalies, LafitE~\citep{yin2023lafite} and DiAD~\citep{he2024diad} via diffusion models,  HVQ-Trans ~\citep{lu2023hierarchical} via vector quantization, ReContrast~\citep{guo2023recontrast} via contrastive cross-reconstruction, CCL~\citep{fan2025salvaging} via class-aware representation learning, DecAD~\citep{wang2025decad} via decoupling anomalies in latent space, ViTAD~\citep{zhang2023exploring} via Vision Transformer (ViT) features, and MambaAD~\citep{he2024mambaad} via state space models.
Despite these efforts, existing unified models still exhibit a notable performance gap relative to SOTA single-class methods, limiting the practical applicability of multi-class UAD.

Our preliminary work, Dinomaly~\citep{guo2025dinomaly}, closed this gap on standard industrial benchmarks, demonstrating that simple components are sufficient to match state-of-the-art single-class performance. The simplicity of its framework makes it easy to reuse and extend. Several recent methods build on Dinomaly or validate their designs within its framework. INP-Former~\citep{luo2025exploring} and HiProtoAD~\citep{li2026hiprotoad} use learnable tokens to extract prototypes for decoder reconstruction. EILA-AD~\citep{liu2026entropy} and ShortcutBreaker~\citep{tang2025shortcutbreaker} refine Linear Attention and Noisy Bottleneck, respectively, to mitigate identity mapping. Dinomaly+~\citep{zhu2025real} and CRR~\citep{wang2025collaborative} add segmentation modules trained using synthetic anomalies, while MVAG-FL~\citep{zhou2025multi} and GLAD~\citep{chen2026see} introduce cross-view attention for multi-view anomaly detection. These extensions primarily target specific multi-class or multi-view settings through additional modules or training stages.

\textbf{Multi-Modal UAD.} Building upon advances in 2D UAD, recent works have incorporated additional modalities beyond conventional 2D RGB data to enhance anomaly detection capabilities. Real-IAD~\citep{wang2024real} extends traditional paradigm to include multiple camera views for each object in real-world manufacturing environments. MVTec3D~\citep{bergmann2021mvtec3d} explores 3D industrial inspection through point cloud data, addressing geometric anomalies that are invisible in 2D imagery. A number of methods were proposed to leverage both RGB and 3D information for UAD~\citep{horwitz2023back,wang2025m3dm,wang2023multimodal}. To further advance multi-modal UAD, Real-IAD $D^3$~\citep{zhu2025real} provides synchronized 2D, 3D, and pseudo-3D photometric stereo data. MulSen-AD~\citep{li2025multi} incorporates infrared (IR) images to exploit thermal information for anomaly detection.

\textbf{Few-Shot UAD.}
 Real-world scenarios often face data scarcity challenges, particularly during system cold-start or when dealing with rare object categories. Early few-shot UAD approaches leverage vision-language models to compensate for scarce visual data. For example, WinCLIP~\citep{jeong2023winclip} utilizes CLIP's pre-trained vision-language representations and introduces window-based prompting strategies to adapt to specific object categories. PromptAD~\citep{li2024promptad} extends this paradigm by designing learnable prompts that capture domain-specific patterns. Meta-learning approaches represent another direction. For example, InCTRL~\citep{zhu2024toward} and MetaUAS~\citep{gao2024metauas} employ meta-learning frameworks to learn transferable detection strategies from the auxiliary datasets and apply them to down-stream datasets. However, both multi-modal and few-shot extensions require specialized architectures and distinct pipelines, lacking a unified framework across diverse scenarios.

\section{Method}
\subsection{Framework Overview}
\noindent The ability to recognize anomalies from what we know is an innate human capability, serving as a vital pathway for us to explore the world. Similarly, we construct a reconstruction-based framework that relies on the epistemic characteristic of artificial neural networks. Dinomaly2 consists of an encoder, a bottleneck, and a reconstruction decoder, as shown in Fig.~\ref{fig2}.

Given an input image $\mathbf{x} \in \mathbb{R}^{H \times W \times C}$, a pre-trained Vision Transformer (ViT) $\mathcal{E}$ with $L$ layers serves as the feature extractor: 
\begin{equation}
\mathbf{F} = \mathcal{E}(\mathbf{x}) = \{\mathbf{f}_1, \mathbf{f}_2, \ldots, \mathbf{f}_L\},
\end{equation}
where $\mathbf{f}_i \in \mathbb{R}^{N \times d}$ represents the feature representation at the $i$-th layer, with $N$ being the number of tokens and $d$ the feature dimension. By default, we use a ViT-Base~\citep{dosovitskiy2020image} architecture with $L=12$ layers. We collect a group of features $\mathbf{F}^*$ from the $L^*$=8 middle layers $\mathcal{M}=\{3,...,10\}$ to capture multi-scale semantic information:
\begin{equation}
\mathbf{F}^* =\{\mathbf{f}_i \mid i \in \mathcal{M}\}= \{\mathbf{f}_3, \mathbf{f}_4, \ldots, \mathbf{f}_{10}\}.
\end{equation}
The bottleneck $\mathcal{B}$ is implemented as a simple multi-layer perceptron (MLP) that processes selected intermediate features:  
\begin{equation}
\mathbf{z} = \mathcal{B}(\mathbf{z}_0)=\mathcal{B}(\sum_{i \in \mathcal{M}} \mathbf{f}_i),
\end{equation}
where $\mathbf{z}$ is the output of $\mathcal{B}$. The decoder $\mathcal{D}$ consists of $L_d$ Transformer layers that reconstruct the encoder's intermediate features: 
\begin{equation}
\hat{\mathbf{F}} = \mathcal{D}(\mathbf{z}) = \{\hat{\mathbf{f}}_{10}, \hat{\mathbf{f}}_9, \ldots, \hat{\mathbf{f}}_3\},
\end{equation}
where $\hat{\mathbf{f}}_i$ represents the reconstructed feature at layer $i$, with $L_d$=8 corresponding to the number of selected encoder layers.  

During training, the model learns to reconstruct normal patterns by optimizing the feature-level reconstruction loss. The primary objective is to minimize the cosine distance between original and reconstructed features: 
\begin{align}
\mathcal{L}_{\text{recon}} = \frac{1}{|\mathcal{M}|} \sum_{i \in \mathcal{M}} {d}_{\text{cos}}(\mathcal{F}(\mathbf{f}_i), \mathcal{F}(\hat{\mathbf{f}}_i)),
\end{align}
where $\mathcal{F}$ denotes the flattening operation and ${d}_{\text{cos}}$ denotes the cosine distance. The model is trained exclusively on normal samples $\mathbf{x}$$\sim$$ \mathcal{X}_{\text{normal}}$, learning to capture the underlying distribution of normal patterns while remaining ignorant of anomalous variations. During inference, the decoder is expected to reconstruct normal regions accurately while producing larger reconstruction errors in anomalous regions, whose patterns were not observed during training.

\textbf{Foundational Representations.}
Recent studies ~\citep{reiss2022anomaly,zhang2023exploring} have demonstrated that self-supervised Vision Transformers (ViTs) ~\citep{zhou2021ibot,oquab2023dinov2,simeoni2025dinov3} provide more robust and universal features for UAD compared to domain-specific ImageNet features.
In this paper, we present the first systematic investigation of the scaling behaviors of vision foundation models in UAD, as illustrated in Fig.~\ref{fig1}(e). Our comprehensive evaluation encompasses model size (Table~\ref{tab:arch}), input resolution (Table~\ref{tab:isize}), and pre-training strategy (Table~\ref{tab:pretrain}), which are detailed in the Experiments section.

\subsection{Challenges and Design Rationale}
\label{sec:design_rationale}

\textbf{Over-generalization.} Prior studies~\citep{you2022unified,zhao2023omnial,guo2023recontrast} attribute the performance degradation of UAD methods trained on diverse multi-class samples to the ``identity mapping'' phenomenon, where inputs are directly copied to outputs. In this work, we reinterpret this issue through the broader perspective of ``over-generalization.'' Generalization ability is a desirable property of neural networks, allowing them to perform well on unseen samples. However, excessive generalization is undesirable in the context of unsupervised anomaly detection based on reconstruction error: even without learning a literal identity mapping, a decoder may extend its reconstruction ability from normal training patterns to unseen anomalous patterns. In multi-class UAD settings, the increasing diversity of images and patterns may exacerbate this problem and weaken reconstruction-based anomaly detection.

\textbf{Contextual ambiguity.} A fundamental challenge in multi-class UAD arises from the context-dependent nature of anomalies: identical visual features can be normal in one context but anomalous in another. For example, a vehicle is normal on a highway but anomalous on a pedestrian walkway, whereas a pedestrian is expected on a sidewalk but becomes anomalous on a highway (Fig.~\ref{fig4}(a)). This contextual ambiguity becomes particularly problematic in unified multi-class models, where the decoder may learn to reconstruct both vehicles and pedestrians in any context (Fig.~\ref{fig4}(b)), reducing its ability to distinguish normal from anomalous patterns.

\textbf{Design philosophy.} Built on foundational ViT features, Dinomaly2 follows a single ``less is more'' design principle to address these two challenges: \textit{less information} through Dropout, \textit{less attention} through Unfocused Linear Attention, \textit{less constraint} through Loose Reconstruction, and \textit{less context invariance} through Context-Aware Recentering. The first three components curb over-generalization by limiting the decoder's ability to reconstruct anomalous inputs; the last one complements them by addressing contextual ambiguity through context-aware reconstruction targets. Together, these simple operations form a reconstruction framework that extends naturally across diverse tasks with minimal task-specific adaptations.

\subsection{Noisy Bottleneck}

A direct solution for identity mapping is to shift ``reconstruction'' to ``restoration''. Specifically, instead of directly reconstructing the normal images or features given normal inputs, perturbations have been introduced as pseudo anomalies, either on input images ~\citep{zavrtanik2021draem,zhang2023destseg} or on features ~\citep{you2022unified,yin2023lafite}. The decoder is still required to restore anomaly-free images or features, formulating a denoising-like framework. However, such methods rely on heuristic and synthetic anomaly generation strategies that may not generalize well across domains, datasets, and methods. 

\textit{“\textbf{Dropout is all you need.}”}

In this work, we turn to leveraging Dropout, a simple yet effective technique originally proposed to mitigate overfitting~\citep{srivastava2014dropout}. In Dinomaly2, Dropout is applied to randomly discard neural activations in an MLP bottleneck. The Noisy Bottleneck $\mathcal{B}$ is implemented as an MLP with $L_b$=3 layers. Given the aggregated features $\mathbf{z}_0 = \sum_{i \in \mathcal{M}} f_i$ from the encoder's layers, $\mathcal{B}$ processes them through sequential transformations:
\begin{equation}
\mathbf{z} = (\mathbf{m}_{L_b} \odot \phi_{L_b}) \circ (\mathbf{m}_{L_b-1} \odot \phi_{L_b-1}) \circ \cdots \circ (\mathbf{m}_1 \odot \phi_1)(\mathbf{z}_0),
\end{equation}
where $\mathbf{m} \sim \text{Bernoulli}(1-p)$ is a random binary mask, $\phi$ is a linear layer (with activation), and $\odot$ denotes element-wise multiplication. 

The role of the Noisy Bottleneck can be explained as introducing \textit{pseudo feature anomaly}. During training, Dropout creates an exponentially large ensemble of $2^{n \cdot L_d}$ sub-networks (where $n$ is the number of units of each linear layer), each observing a different ``corrupted'' version of the features. This provides a parameter-free, domain-agnostic corruption mechanism, forcing the decoder to map corrupted inputs back to the most likely normal patterns.

The role of the Noisy Bottleneck can also be interpreted through the \textit{information bottleneck} (IB) principle ~\citep{achille2018information,tishby2015deep}. The UAD task imposes dual constraints that distinguish between normal and anomalous input domains. The IB objective can be formulated as:
\begin{equation}
\max_{\mathcal{B},\mathcal{D}} I(\mathbf{z}; \hat{\mathbf{F}} | \mathbf{x} \in \mathcal{X}_{\text{normal}}) - \beta I(\mathbf{z}; \mathbf{F}^* | \mathbf{x} \in \mathcal{X}_{\text{all}})
\end{equation}
where $\mathcal{X}_{\text{normal}}$ denotes the normal domain, $\mathcal{X}_{\text{all}}=\mathcal{X}_{\text{normal}} \cup \mathcal{X}_{\text{ano}}$ denotes the complete input space, $I(\cdot;\cdot)$ denotes mutual information, and $\beta$ controls the compression-reconstruction trade-off. While direct access to $\mathcal{X}_{\text{ano}}$ is unavailable, Dropout provides an implicit constraint on $ I(\mathbf{z}; \mathbf{F}^* | \mathbf{x} \in \mathcal{X}_{\text{ano}})$. The key insight is that Dropout does not uniformly reduce $\mathcal{B}$'s capacity; instead, it enforces a redundant encoding scheme specifically tailored to the statistical structure of $\mathcal{X}_{\text{normal}}$. Formally, the learned encoding maximizes:
\begin{equation}
\mathbb{E}_{\mathbf{M}} \left[ I(\mathcal{B}(\mathbf{F}^*; \mathbf{M}); \hat{\mathbf{F}} | \mathbf{x} \in \mathcal{X}_{\text{normal}}) \right],
\end{equation}
where $\mathbf{M} = \{\mathbf{m}_1, \ldots, \mathbf{m}_{L_b}\}$ denotes the collection of independent dropout masks. The expectation is taken over all possible realizations of dropout masks across all layers. Since the overall capacity of $\mathcal{B}$ is constrained, $I(\mathbf{z}; \mathbf{F}^* | \mathbf{x} \in \mathcal{X}_{\text{ano}})$ is implicitly limited by the redundant encoding of observed normal patterns and the inefficient encoding of unseen anomalous patterns.

\subsection{Unfocused Linear Attention}

\textbf{Softmax Attention} ~\citep{vaswani2017attention} is the cornerstone of Transformer architectures, enabling models to dynamically capture relationships between different parts of the input. However, in the context of multi-class anomaly detection, the very strength of attention becomes a liability. 

\begin{figure}[!t]
\centerline{\includegraphics[width=0.99\linewidth]{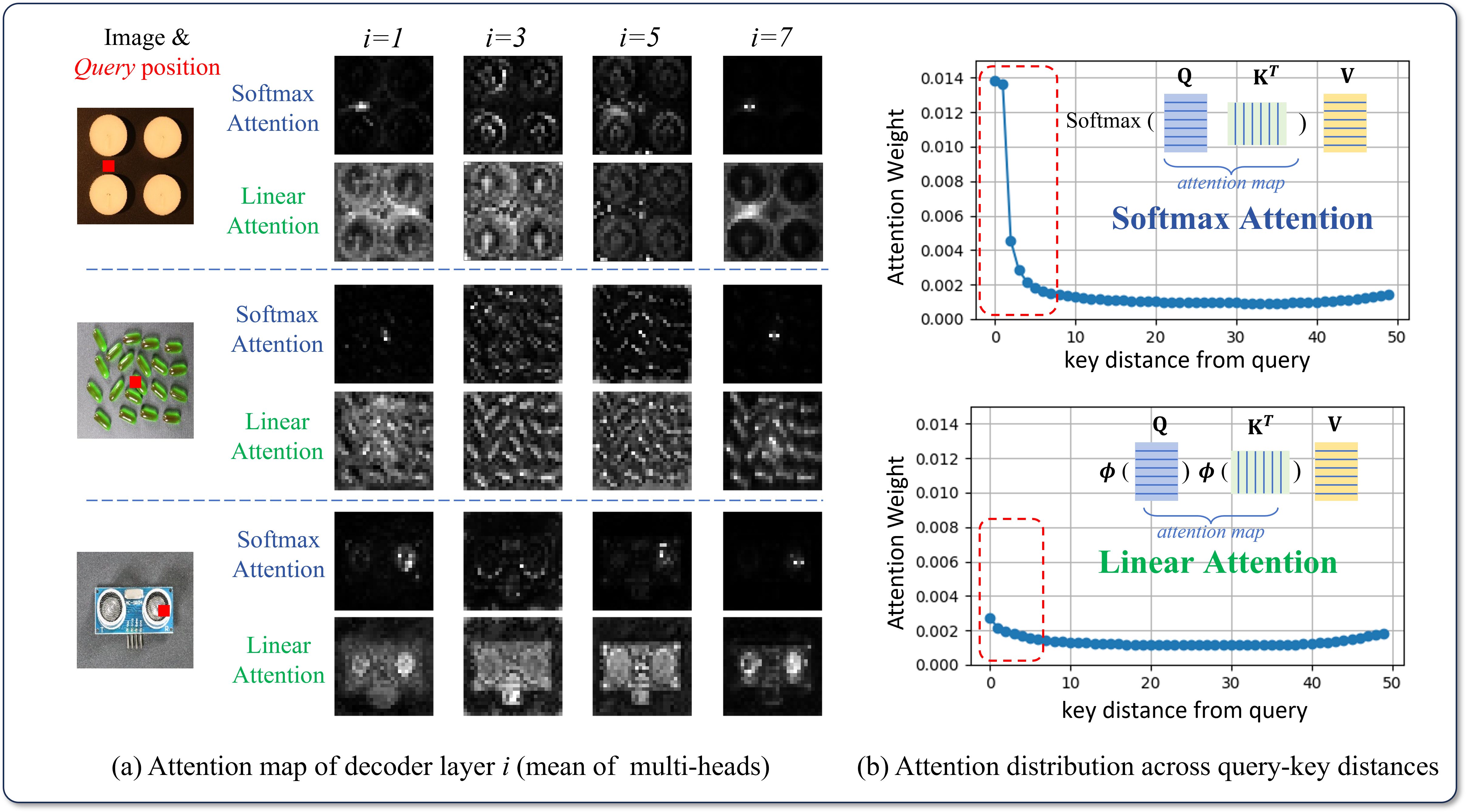}}
\caption{Softmax Attention \textit{vs.} Linear Attention. (a) Attention maps and (b) attention distributions illustrate Softmax Attention's tendency to focus on local regions, in contrast to Linear Attention's distributed attention pattern.}
\label{fig3}
\end{figure}

\textbf{The Focusing Paradox.} Formally, given an input sequence $\mathbf{f} \in \mathbb{R}^{N \times d}$ with length $N$, attention first transforms it into three matrices: the query matrix $\mathbf{Q} \in \mathbb{R}^{N \times d}$, the key matrix $\mathbf{K} \in \mathbb{R}^{N \times d}$, and the value matrix $\mathbf{V} \in \mathbb{R}^{N \times d}$:
\begin{equation}
\mathbf{Q} = \mathbf{f} \mathbf{W}^Q ~,
\mathbf{K} = \mathbf{f} \mathbf{W}^K ~,
\mathbf{V} = \mathbf{f} \mathbf{W}^V ~,
\end{equation}
where $\mathbf{W}^Q, \mathbf{W}^K, \mathbf{W}^V \in \mathbb{R}^{d \times d}$ are learnable parameters. Softmax Attention (SA) computes the attention map from query--key similarities and produces the following output:  \footnote{The full form of Attention is $\text{Softmax}(\frac{\mathbf{Q}\mathbf{K}^T}{\sqrt{d}}) \mathbf{V}$. The constant denominator and multi-head mechanism are omitted for narrative simplicity.}
\begin{equation}
\text{Attention}(\mathbf{Q}, \mathbf{K}, \mathbf{V})= \text{Softmax}(\mathbf{Q}\mathbf{K}^T) \mathbf{V} ~.
\end{equation}
In UAD, previous methods ~\citep{you2022unified,lu2023hierarchical} replace convolutions with attention, because convolutional networks are prone to learning identity mappings. Nevertheless, the softmax mechanism inherently encourages the model to concentrate on the most relevant features, forming identity mapping by only concentrating on corresponding input locations (Fig.~\ref{fig3}).

\textit{“\textbf{One man's poison is another man's meat.}”}

\textbf{Linear Attention as a Feature, Not a Bug.} Linear Attention (LA) was proposed as an alternative to reduce the computational complexity of vanilla Softmax Attention with respect to the number of tokens ~\citep{katharopoulos2020transformers}. By replacing the softmax operation with a simple activation function $\phi( \cdot )$ (usually $\phi(x) = \text{elu}(x)+1$), the computation order changes from $(\mathbf{Q} \mathbf{K}^T)\mathbf{V}$ to $\mathbf{Q} (\mathbf{K}^T \mathbf{V})$. Formally, LA is given as:
\begin{equation}
\text{LA}(\mathbf{Q}, \mathbf{K}, \mathbf{V})= (\phi(\mathbf{Q}) \phi(\mathbf{K}^T)) \mathbf{V} = \phi(\mathbf{Q})(\phi(\mathbf{K}^T) \mathbf{V}) ~,
\end{equation}
where the computational complexity is reduced from $\mathcal{O}(N^2d)$ to $\mathcal{O}(Nd^2)$ . The trade-off between complexity and expressiveness is a dilemma. Previous studies ~\citep{han2023flatten} attribute LA's performance degradation on supervised tasks to its limited ability to focus on query-relevant regions, such as foreground regions or neighboring tokens. This property, however, is useful for our reconstruction decoder. 

In Dinomaly2, we leverage LA's inherent inability to focus as a desirable property for anomaly detection. To examine how attention propagates information, we train two variants of Dinomaly using vanilla SA or LA as the spatial mixer in the decoder and visualize their attention maps. As shown in Fig.~\ref{fig3}(a), SA tends to focus on the exact region of the query, while LA spreads its attention across the entire image. From a frequency domain view, LA acts as a low-pass filter ~\citep{wi2025learning} that cannot selectively amplify local high-frequency details (Fig.~\ref{fig3}(b)), compelling the network to reconstruct based on learned global patterns. This diffusive propagation prevents identity mapping in anomaly detection.

\subsection{Context-Aware Recentering}

\begin{figure}[!t]
\centerline{\includegraphics[width=0.99\linewidth]{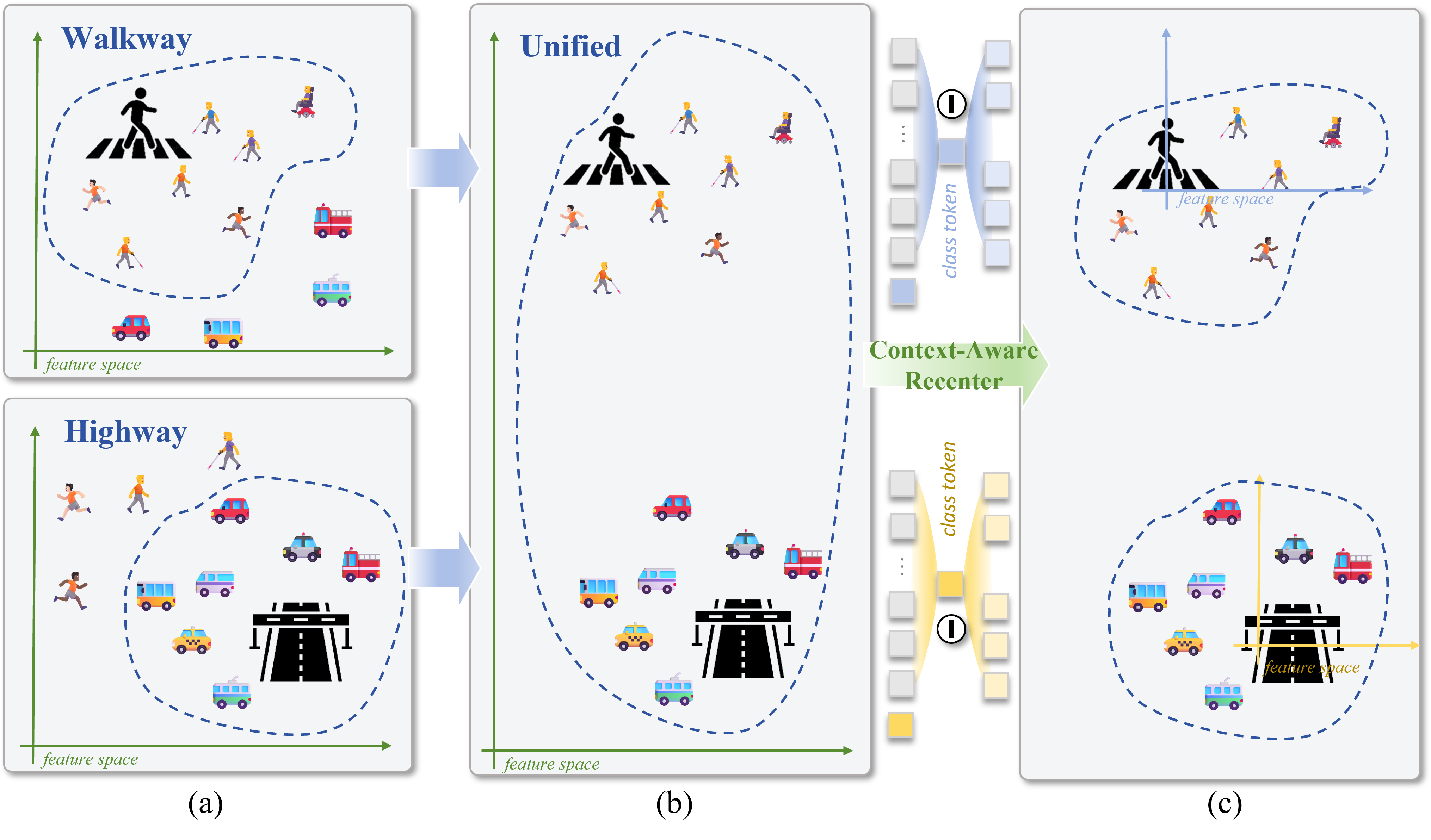}}
\caption{Context-Aware Recentering and a toy example. (a) Separate detectors for sidewalk and highway show how identical objects (pedestrians/vehicles) may receive opposite anomaly labels depending on context.  (b) A unified detector faces contextual ambiguity without proper conditioning.  (c) Our Context-Aware Recentering uses CLS tokens as reference anchors to map patch features into class-specific coordinate systems, resolving multi-class confusion through simple subtraction. }
\label{fig4}
\end{figure}

In multi-class UAD, the normality of a local pattern depends on its surrounding context. Consequently, the same visual pattern may be normal in one image but anomalous in another, making context-independent reconstruction ambiguous (Fig.~\ref{fig4}).

\textit{"\textbf{Beauty is in the eye of the beholder."}}
x
Motivated by this observation, we introduce Context-Aware Recentering (CAR), a simple yet effective mechanism that conditions feature reconstruction on class-specific context. The key insight is to leverage the CLS token from Vision Transformers as a contextual anchor that encodes class-specific normal patterns.
Formally, let $\mathbf{f}_i \in \mathbb{R}^{(N+1) \times d}$ denote the feature representation at encoder layer $i$, where the first token $\mathbf{f}_i^{cls} \in \mathbb{R}^{d}$ represents the CLS token and the remaining $\mathbf{f}_i^{patch} \in \mathbb{R}^{N \times d}$ represent patch tokens. Instead of reconstructing the patch features directly, we reconstruct the recentered features:
\begin{equation} \tilde{\mathbf{f}}_i^{patch} = \mathbf{f}_i^{patch} - \mathbf{f}_i^{cls} \otimes \mathbf{1}_N \end{equation}
where $\otimes$ denotes the outer product and $\mathbf{1}_N$ is a vector of ones. This subtraction expresses each patch feature relative to the image-specific CLS context (Fig.~\ref{fig4}(c)), allowing the same local pattern to have different reconstruction targets in different contexts. The decoder reconstructs these context-aware targets without requiring class labels.

\begin{figure*}[!t]
\centerline{\includegraphics[width=0.99\textwidth]{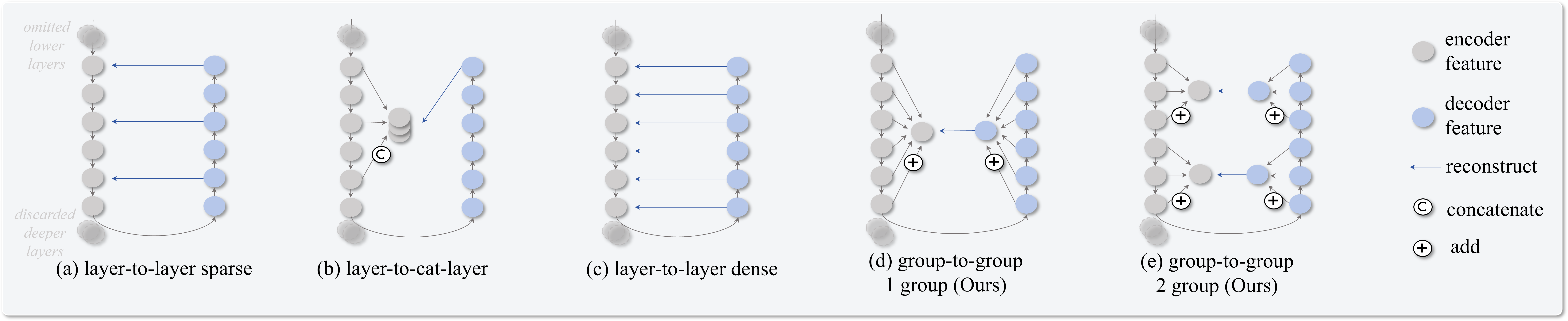}}
\setlength{\abovecaptionskip}{2pt} 
\caption{Schemes of reconstruction constraint. (a) Layer-to-layer (sparse). (b) Layer-to-cat-layer. (c) Layer-to-layer (dense).  (d) Loose group-to-group, 1-group (Ours). (e) Loose group-to-group, 2-group (Ours). }
\label{fig5}
\end{figure*}

\subsection{Loose Reconstruction}

A critical challenge in multi-class UAD lies in the reconstruction objective design. Prior feature-reconstruction UAD methods ~\citep{deng2022anomaly,salehi2021multiresolution,zhang2023exploring} typically follow knowledge distillation paradigms ~\citep{hinton2015distilling}, where decoder layers are trained to precisely mimic corresponding encoder layers. While this layer-to-layer supervision provides strong learning signals, it may also make the decoder overly capable of reconstructing anomalous patterns that were not encountered during training.
We propose Loose Reconstruction, which deliberately relaxes both structural constraints and optimization objectives to prevent over-generalization in multi-class settings.

\textit{"\textbf{The tighter you squeeze, the less you have.}"}

\textbf{Loose Constraint}. Instead of enforcing strict layer-to-layer correspondence between encoder and decoder features (Fig.~\ref{fig5}(a-c)), we group multiple layers into semantic clusters:
\begin{equation} 
\mathbf{g}_k = \sum_{i \in \mathcal{S}_k} \mathbf{f}_i, k \in \mathcal{G}
\end{equation}
where $\mathcal{S}_k$ denotes a group of layers. All layers may be grouped into a single cluster, so that $\mathcal{G}=\{1\}$, $\mathcal{S}_1 = \{3,4,...,10\}$ (Fig.~\ref{fig5}(d)). Given the hierarchical nature of Vision Transformers, where shallow layers capture low-level visual features and deeper layers encode high-level semantics, we aggregate features from multiple encoder layers: $\mathcal{G}=\{1,2\}$, $\mathcal{S}_1 = \{3,4,5,6\}$ and $\mathcal{S}_2 = \{7,8,9,10\}$ represent shallow and deep layer groups respectively (Fig.~\ref{fig5}(e)). The decoder then reconstructs these grouped representations rather than individual layers: 
\begin{equation} 
\hat{\mathbf{g}}_k = \sum_{j \in \mathcal{S}_k'} \hat{\mathbf{f}}_j, k \in \mathcal{G}.
\end{equation}
This scheme relaxes layer-to-layer correspondence and gives the decoder more freedom to deviate from the encoder when processing unseen patterns. 

\textbf{Loose Loss}. Beyond structural relaxation, we further introduce a selective optimization strategy that dynamically adjusts gradient flow based on reconstruction quality. Here, we directly follow our previous work ~\citep{guo2023recontrast}, adopting a hard-mining global cosine loss that selectively reduces gradients for well-reconstructed regions:
\begin{equation}
\mathcal{L}_{\text{loose}} = \frac{1}{|\mathcal{G}|} \sum_{k \in \mathcal{G}} {d}_{\text{cos}}(\mathcal{F}(\mathbf{g}_k), \mathcal{F}(\hat{\mathbf{g}}'_k)),
\end{equation}
where $\hat{\mathbf{g}}'_k$ undergoes selective gradient modulation: 
\begin{equation} 
\hat{\mathbf{g}}'_k(n) = \begin{cases} \text{sg}_{0.1}(\hat{\mathbf{g}}_k(n)), 
& \text{if } {d}_{\text{cos}}(\mathbf{g}_k(n), \hat{\mathbf{g}}_k(n)) > \tau_k \\ \hat{\mathbf{g}}_k(n), & \text{otherwise}
\end{cases}, \end{equation}
where $\text{sg}_{0.1}(\cdot)$ reduces gradients to 10\% of their original magnitude\footnote{Complete gradient stopping occasionally causes training instability.}, and $\tau_k$ represents the $k$-th percentile threshold within each batch (90\% by default). The combination of loose constraints and loose loss creates a reconstruction objective that is deliberately imperfect—capable of capturing normal patterns while maintaining sufficient reconstruction error on anomalous regions.

\textbf{Anomaly Scoring.} Finally, the reconstruction errors serve as the foundation for anomaly localization and detection. Given encoder feature groups $\mathbf{g}_k$ and their reconstructed counterparts $\hat{\mathbf{g}}_k$, we compute token-wise cosine similarity to generate anomaly maps:
\begin{align}
\mathbf{A} = \frac{1}{|\mathcal{G}|} \sum_{k \in \mathcal{G}} \mathbf{A}_k,
\end{align}
\begin{align}
\mathbf{A}_k(n) = 1 - \frac{\mathbf{g}_k(n) \cdot \hat{\mathbf{g}}_k(n)}{||\mathbf{g}_k(n)||_2 \cdot ||\hat{\mathbf{g}}_k(n)||_2},
\end{align}
where $n$ indexes the spatial tokens. For image-level anomaly detection, we compute the anomaly score as the mean of the top z\% (default 1\%) most anomalous pixels:
\begin{align}
s_{image} = \text{mean}(\text{top}_{z\%}(\mathbf{A})).
\end{align}

\subsection{Seamless Extension Beyond Plain 2D UAD}

The minimalist philosophy of Dinomaly2 naturally extends beyond 2D images to diverse data modalities. We demonstrate that lightweight task-specific adaptations (post-processing, augmentation, \textit{etc.}) of Dinomaly2 achieve state-of-the-art results without introducing complex fusion mechanisms or modality-specific architectures.

\textit{"\textbf{Entia non sunt multiplicanda praeter necessitatem.}"} 

\textbf{Multi-View Inspection.} Industrial inspection often requires examining objects from multiple viewpoints. For multi-view datasets such as Real-IAD~\citep{wang2024real} and MANTA~\citep{fan2025manta}, each object is captured from $V$=5 camera angles. We retain the same training procedure and train jointly on all views from all classes. During inference, anomaly maps from all views are concatenated, and the object-level anomaly score is computed as the mean of the top $z\%\cdot{V}$ most anomalous pixels across the concatenated anomaly map:
\begin{align}
s_{object} = \text{mean}(\text{top}_{z\%\cdot{V}}(\bigcup_{v=1}^{V} \mathbf{A}_v))
\end{align}
where $\mathbf{A}_v$ denotes the anomaly map from view $v$. This straightforward aggregation strategy consolidates multi-view information without requiring view-specific training or complex inter-view interaction.

\textbf{Aligned Multi-Modality.} For the datasets combining RGB images with pixel-aligned 3D point cloud data (\textit{e.g.}, MVTec3D~\citep{bergmann2021mvtec3d}), we adopt a minimal fusion strategy. Following~\citep{rudolph2023asymmetric}, we first project the 3D point cloud to obtain a depth map. The RGB image and depth map are processed independently by the same pre-trained ViT encoder, yielding feature representations $\mathbf{F}^*_{rgb}$ and $\mathbf{F}^*_{depth}$ respectively. The multi-modal encoder representation is obtained through element-wise averaging:
\begin{align}
\mathbf{F}^{*} = \frac{1}{2}(\mathbf{F}_{rgb}^* + \mathbf{F}_{depth}^*).
\end{align}
The subsequent bottleneck, decoder, and reconstruction processes remain identical to the standard framework. This naive fusion, requiring no additional parameters or architectural modifications, allows Dinomaly2 to leverage complementary geometric and appearance information. 

\textbf{Non-aligned Multi-Modality.} In some scenarios, multi-modal information is acquired without spatial alignment, where images of different modalities are acquired from different camera angles. Here, we adopt a straightforward extension of our multi-view inspection strategy: each modality is treated as an independent view. For example, in MulSen-AD~\citep{li2025multi}, RGB and infrared (IR) images are processed as separate views within our unified framework. The final object-level anomaly score is computed as the sum of the individual image-level anomaly scores from both modalities.

\textbf{Few-Shot Anomaly Detection.} Real-world UAD scenarios often encounter data scarcity, where only limited normal samples are available. Dinomaly2's minimalist design naturally adapts to the multi-class few-shot setting without any architectural modifications. Our only adaptation is standard data augmentation during training, including random flipping, rotation, and translation. These elementary augmentations expand the limited normal sample space without introducing domain-specific assumptions (text-prompts)~\citep{jeong2023winclip} or complex meta-learning strategies~\citep{gao2024metauas} employed by specialized few-shot UAD methods.

\section{Experiments}
\subsection{Experimental Settings}

\begin{table}[t]
	\centering
	\caption{Statistics of 15 UAD datasets.}
	\label{tab:anomaly_datasets}
	\footnotesize
	\setlength{\tabcolsep}{1pt}
	\begin{tabular}{l|cccc}
		\hline
		\textbf{Dataset} & \textbf{Mod}  & \#C & {\tiny\#Train/Test(good/bad)} & {\tiny Train Iters} \\
		\hline
		MVTec-AD  &RGB & 15 & 3,629/467/1,258 &4$\times$$10^4$  \\
		VisA &RGB & 12 & 8,659/962/1,200 &4$\times$$10^4$ \\
		MPDD &RGB & 6 & 888/176/282 &2$\times$$10^4$   \\
		BTAD &RGB & 3 & 1,799/451/290 &2$\times$$10^4$   \\
		Real-IAD  & MV  & 30 & 36,465/63,256/51,329 &1$\times$$10^5$   \\
		MANTA & MV  & 38 & 21,483/33,560/11,025 &1$\times$$10^5$   \\
		\makecell[l]{Real-IAD-Variety} & MV & 160 & 19,955/23,185/155,810 & $1\times10^5$ \\
		ADNet & RGB & 380 & 116,192/19,791/60,311 & $1\times10^5$ \\
		MVTec LOCO  &RGB & 5 & 1,772/575/993 &$1\times10^4$  \\
		MVTec3D  &RGB-3D & 10 & 2,656/249/948 &4$\times$$10^4$  \\
		MulSen-AD  &RGB-IR & 15 & 1,391/153/491 &2$\times$$10^4$   \\
		Uni-Medical &Gray & 3 & 13,339/2,514/4,499 &4$\times$$10^4$   \\
		ApoCell &Gray & 1 & 2,704/1,597/1,016 &1$\times$$10^4$   \\
		MIAD &RGB & 7 & 70,000/17,500/17,500 &1$\times$$10^5$   \\
		Drone-Ano. &RGB & 7 & 51,635/19,724/16,119 &1$\times$$10^4$   \\
		\bottomrule
	\end{tabular}
\end{table}

We conduct a comprehensive evaluation across a wide range of benchmarks, referred to as our UAD Dataset Marathon (Table~\ref{tab:anomaly_datasets}). Specificially, \textbf{MVTec-AD} ~\citep{bergmann2019mvtec} is the most widely-used industrial anomaly detection benchmark, containing 5 texture classes and 10 object classes. 
\textbf{VisA} ~\citep{zou2022spot} focuses on complex industrial products with challenging anomaly patterns, \textit{e.g.}, multiple objects in one image.
\textbf{MPDD} ~\citep{jezek2021deep} targets visual defect detection in metal parts manufacturing with real-world industrial conditions.
\textbf{BTAD} ~\citep{mishra2021btad} contains real-world industrial products, with anomalous images inadvertently mixed into BTAD-02's normal training split without being identified as anomalies~\citep{jiang2022softpatch}. We retain the official split for comparability.

\textbf{Real-IAD} ~\citep{wang2024real} is a large multi-view dataset containing 30 categories, each with 5 camera views. We follow the official splitting. \textbf{MANTA}~\citep{fan2025manta} is a large-scale multi-view dataset for tiny objects (e.g. various seeds and beans). We use the official MANTA-Tiny split. \textbf{Real-IAD Variety}~\citep{zhu2026real} expands industrial inspection to 160 fine-grained product varieties, providing a more challenging large-scale multi-class and multi-view benchmark. \textbf{ADNet}~\citep{ling2026adnet} spans 380 real-world categories across multiple domains, substantially increasing category diversity and scale. \textbf{MVTec LOCO}~\citep{bergmann2022beyond} contains five industrial categories with both structural defects and logical anomalies. \textbf{MVTec3D} ~\citep{bergmann2021mvtec3d} contains point clouds scanned by industrial 3D sensors with paired 2D RGB images from 10 categories. 3D point clouds are pre-processed to depth maps following ~\citep{rudolph2023asymmetric}. \textbf{MulSen-AD} ~\citep{li2025multi} is a multi-sensor dataset unifying data from RGB cameras, laser scanners, and lock-in infrared thermography. We utilize RGB and IR images.

\textbf{Uni-Medical} ~\citep{zhang2024ader} is a medical UAD dataset consisting of 2D slices of brain CT, liver CT, and retinal OCT, extracted from ~\citep{bao2024bmad}. \textbf{ApoCell} consists of confocal fluorescence  microscopy images from \textit{C. elegans} embryos, derived from ~\citep{guan2025cell}. Normal images contain only viable cells, while anomalous images include cells undergoing apoptosis. \textbf{MIAD} ~\citep{bao2023miad} consists of outdoor images with uncontrolled factors (viewpoint, background, etc.), focusing on maintenance inspection to keep equipment in optimum working condition. \textbf{Drone-Anomaly} ~\citep{jin2022anomaly} is a drone-based dataset for abnormal event surveillance captured by UAV cameras. This dataset contains 7 distinct scenes including highway, crossroad, bike roundabout, etc.

\begin{table*}[t]
  \centering
  \caption{{Multi-class} UAD performance on conventional 2D industrial datasets (\%). \dag~denotes methods originally designed for single-class UAD. \underline{Underlines} represent runner-up performance excluding Dinomaly variants.}
   \resizebox{0.99\textwidth}{!}{
    \begin{tabular}{l|ccccccc|ccccccc}
    \toprule
    \multirow{3}[3]{*}{Method} & \multicolumn{7}{c|}{\textbf{MVTec-AD}~\citep{bergmann2019mvtec} (15 classes)} & \multicolumn{7}{c}{\textbf{VisA}~\citep{zou2022spot} (12 classes)} \\
% \cmidrule(r){2-8} \cmidrule(l){9-15}
    & \multicolumn{3}{c}{Image-level} & \multicolumn{4}{c|}{Pixel-level} & \multicolumn{3}{c}{Image-level} & \multicolumn{4}{c}{Pixel-level} \\
\cmidrule(r){2-4} \cmidrule(lr){5-8} \cmidrule(r){9-11} \cmidrule(l){12-15}
    & AUROC & AP & $F_1$-max & AUROC & AP & $F_1$-max & AUPRO & AUROC & AP & $F_1$-max & AUROC & AP & $F_1$-max & AUPRO \\
\hline
    RD4AD\dag~\citep{deng2022anomaly} & 94.6 & 96.5 & 95.2 & 96.1 & 48.6 & 53.8 & 91.1 & 92.4 & 92.4 & 89.6 & 98.1 & 38.0 & 42.6 & 91.8 \\
    SimpleNet\dag~\citep{liu2023simplenet} & 95.3 & 98.4 & 95.8 & 96.9 & 45.9 & 49.7 & 86.5 & 87.2 & 87.0 & 81.8 & 96.8 & 34.7 & 37.8 & 81.4 \\
    DeSTSeg\dag~\citep{zhang2023destseg} & 89.2 & 95.5 & 91.6 & 93.1 & 54.3 & 50.9 & 64.8 & 88.9 & 89.0 & 85.2 & 96.1 & 39.6 & 43.4 & 67.4 \\
    UniAD~\citep{you2022unified} & 96.5 & 98.8 & 96.2 & 96.8 & 43.4 & 49.5 & 90.7 & 88.8 & 90.8 & 85.8 & 98.3 & 33.7 & 39.0 & 85.5 \\
    ReContrast~\citep{guo2023recontrast} & 98.3 & 99.4 & 97.6 & 97.1 & \underline{60.2} & \underline{61.5} & \underline{93.2} & \underline{95.5} & \underline{96.4} & \underline{92.0} & \underline{98.5} & \underline{47.9} & \underline{50.6} & \underline{91.9} \\
    DiAD~\citep{he2024diad} & 97.2 & 99.0 & 96.5 & 96.8 & 52.6 & 55.5 & 90.7 & 86.8 & 88.3 & 85.1 & 96.0 & 26.1 & 33.0 & 75.2 \\
    ViTAD~\citep{zhang2023exploring} & 98.3 & 99.4 & 97.3 & \underline{97.7} & 55.3 & 58.7 & 91.4 & 90.5 & 91.7 & 86.3 & 98.2 & 36.6 & 41.1 & 85.1 \\
    MambaAD~\citep{he2024mambaad} & \underline{98.6} & \underline{99.6} & \underline{97.8} & \underline{97.7} & 56.3 & 59.2 & 93.1 & 94.3 & 94.5 & 89.4 & \underline{98.5} & 39.4 & 44.0 & 91.0 \\
\rowcolor{ours!50}  Dinomaly-B~\citep{guo2025dinomaly} & 99.6 & 99.8 & 99.0 & 98.4 & 69.3 & 69.2 & 94.8 & 98.7 & 98.9 & 96.2 & 98.7 & 53.2 & 55.7 & 94.5 \\
\rowcolor{ours}  \textbf{Dinomaly2-B} & 99.8 & 99.9 & 99.3 & 98.4 & 69.3 & 68.9 & 94.8 & 99.2 & 99.3 & 97.0 & 99.1 & 52.9 & 56.4 & \textbf{95.5} \\
\rowcolor{ours}  \textbf{Dinomaly2-L} & \textbf{99.9} & \textbf{100.0} & \textbf{99.5} & \textbf{98.6} & \textbf{70.3} & \textbf{69.9} & \textbf{95.1} & \textbf{99.3} & \textbf{99.4} & \textbf{97.2} & \textbf{99.2} & \textbf{54.7} & \textbf{57.1} & 95.4 \\

\hline\hline

    & \multicolumn{7}{c|}{\textbf{MPDD}~\citep{jezek2021deep} (6 classes)} & \multicolumn{7}{c}{\textbf{BTAD}~\citep{mishra2021btad} (3 classes)} \\
% \cmidrule(r){2-8} \cmidrule(l){9-15}
    & \multicolumn{3}{c}{Image-level} & \multicolumn{4}{c|}{Pixel-level} & \multicolumn{3}{c}{Image-level} & \multicolumn{4}{c}{Pixel-level} \\
\cmidrule(r){2-4} \cmidrule(lr){5-8} \cmidrule(r){9-11} \cmidrule(l){12-15}
    & AUROC & AP & $F_1$-max & AUROC & AP & $F_1$-max & AUPRO & AUROC & AP & $F_1$-max & AUROC & AP & $F_1$-max & AUPRO \\
\hline
    RD4AD\dag~\citep{deng2022anomaly} & 90.3 & 92.8 & 90.5 & \underline{98.3} & 39.6 & 40.6 & 95.2 & 94.1 & 96.8 & 93.8 & \underline{98.0} & 57.1 & \underline{58.0} & \textbf{79.9} \\
    SimpleNet\dag~\citep{liu2023simplenet} & 90.6 & \underline{94.1} & 89.7 & 97.1 & 33.6 & 35.7 & 90.0 & 94.0 & 97.9 & 93.9 & 96.2 & 41.0 & 43.7 & 69.6 \\
    DeSTSeg\dag~\citep{zhang2023destseg} & \underline{92.6} & 91.8 & \underline{92.8} & 90.8 & 30.6 & 32.9 & 78.3 & 93.5 & 96.7 & 93.8 & 94.8 & 39.1 & 38.5 & 72.9 \\
    UniAD~\citep{you2022unified} & 80.1 & 83.2 & 85.1 & 95.4 & 19.0 & 25.6 & 83.8 & 94.5 & 98.4 & 94.9 & 97.4 & 52.4 & 55.5 & \underline{78.9} \\
    ReContrast~\citep{guo2023recontrast} & 90.9 & 94.5 & 91.0 & 98.8 & \underline{46.8} & \underline{48.9} & \underline{95.6} & \underline{94.8} & \underline{98.9} & \underline{95.5} & 97.2 & 56.8 & 56.2 & 76.4 \\
    DiAD~\citep{he2024diad} & 85.8 & 89.2 & 86.5 & 91.4 & 15.3 & 19.2 & 66.1 & 90.2 & 88.3 & 92.6 & 91.9 & 20.5 & 27.0 & 70.3 \\
    ViTAD~\citep{zhang2023exploring} & 87.4 & 90.8 & 87.0 & 97.8 & {44.1} & {46.4} & \underline{95.3} & 94.0 & 97.0 & 93.7 & 97.6 & \underline{58.3} & 56.5 & 72.8 \\
    MambaAD~\citep{he2024mambaad} & 89.2 & 93.1 & 90.3 & 97.7 & 33.5 & 38.6 & 92.8 & 92.9 & 96.2 & 93.0 & 97.6 & 51.2 & 55.1 & 77.3 \\

\rowcolor{ours!50}  Dinomaly-B~\citep{guo2025dinomaly} & 97.2 & 98.4 & 96.0 & 99.1 & 59.5 & 59.4 & 96.6 & 95.4 & 98.4 & 95.6 & 97.8 & 70.1 & 68.0 & 76.5 \\
\rowcolor{ours}  \textbf{Dinomaly2-B} & 97.6 & 98.6 & 96.6 & 99.2 & 63.5 & 62.6 & 96.8 & {97.4} & {98.9} & 97.0 & 98.1 & 73.5 & \textbf{68.4} & \textbf{79.9} \\
\rowcolor{ours}  \textbf{Dinomaly2-L} & \textbf{99.0} & \textbf{99.4} & \textbf{97.8} & \textbf{99.3} & \textbf{64.6} & \textbf{62.8} & \textbf{97.1} & \textbf{97.5} & \textbf{99.2} & \textbf{97.4} & \textbf{98.2} & \textbf{73.8} & {68.3} & 79.8 \\
        
    \bottomrule
    \end{tabular}%
    }
  \label{tab:mvtec}%
\end{table*}%

\begin{table*}[t]
  \centering
  \tiny
  \setlength{\tabcolsep}{3pt}
  \renewcommand{\arraystretch}{1.2}
  \caption{Statistical robustness on MVTec-AD and VisA over five random seeds (\%). Results are reported as $\mathrm{mean}_{\pm\mathrm{std}}^{p-value}$, where the superscript is the two-sided paired $t$-test $p$-value against Dinomaly2-B. $p>0.05$ is shown in yellow. }
  \resizebox{\textwidth}{!}{%
  \begin{tabular}{ll|ccc|cccc}
    \toprule
    Dataset & Method & \multicolumn{3}{c}{Image-level} & \multicolumn{4}{c}{Pixel-level} \\
    & & AUROC & AP & $F_1$-max & AUROC & AP & $F_1$-max & AUPRO \\
    \midrule
    \multirow{4}{*}{\textbf{MVTec-AD}} & \textbf{Dinomaly2-B}
    & $\mathbf{99.80}_{\pm0.02}$ & $\mathbf{99.87}_{\pm0.03}$ & $\mathbf{99.36}_{\pm0.02}$ & $\mathbf{98.34}_{\pm0.01}$ & $\mathbf{69.18}_{\pm0.31}$ & $68.75_{\pm0.22}$ & $\mathbf{94.79}_{\pm0.11}$ \\
    & Dinomaly-B
    & $99.61_{\pm0.03}^{p<0.001}$ & $99.78_{\pm0.01}^{p<0.01}$ & $99.07_{\pm0.10}^{p<0.01}$ & $98.33_{\pm0.02}^{\textcolor{yellow!50!black}{p=0.178}}$ & $69.07_{\pm0.24}^{\textcolor{yellow!50!black}{p=0.358}}$ & $\mathbf{68.93}_{\pm0.22}^{p<0.01}$ & $94.62_{\pm0.07}^{\textcolor{yellow!50!black}{p=0.075}}$ \\
    & ViTAD
    & $98.25_{\pm0.07}^{p<0.001}$ & $99.31_{\pm0.05}^{p<0.001}$ & $97.46_{\pm0.09}^{p<0.001}$ & $97.46_{\pm0.02}^{p<0.001}$ & $58.58_{\pm0.28}^{p<0.001}$ & $61.02_{\pm0.16}^{p<0.001}$ & $91.55_{\pm0.07}^{p<0.001}$ \\
    & MambaAD
    & $98.41_{\pm0.09}^{p<0.001}$ & $99.51_{\pm0.03}^{p<0.001}$ & $97.72_{\pm0.13}^{p<0.001}$ & $97.55_{\pm0.07}^{p<0.001}$ & $56.04_{\pm0.07}^{p<0.001}$ & $58.67_{\pm0.13}^{p<0.001}$ & $93.56_{\pm0.06}^{p<0.001}$ \\
    \midrule
    \multirow{4}{*}{\textbf{VisA}} & \textbf{Dinomaly2-B}
    & $\mathbf{99.15}_{\pm0.06}$ & $\mathbf{99.30}_{\pm0.05}$ & $\mathbf{96.82}_{\pm0.14}$ & $\mathbf{99.05}_{\pm0.01}$ & $\mathbf{53.23}_{\pm0.46}$ & $\mathbf{56.43}_{\pm0.26}$ & $\mathbf{95.42}_{\pm0.06}$ \\
    & Dinomaly-B
    & $98.67_{\pm0.07}^{p<0.001}$ & $98.83_{\pm0.08}^{p<0.001}$ & $96.10_{\pm0.15}^{p<0.01}$ & $98.63_{\pm0.07}^{p<0.001}$ & $52.98_{\pm0.54}^{p<0.01}$ & $55.39_{\pm0.21}^{p<0.01}$ & $94.32_{\pm0.11}^{p<0.001}$ \\
    & ViTAD
    & $90.78_{\pm0.17}^{p<0.001}$ & $91.75_{\pm0.12}^{p<0.001}$ & $87.23_{\pm0.14}^{p<0.001}$ & $97.52_{\pm0.02}^{p<0.001}$ & $39.35_{\pm0.12}^{p<0.001}$ & $44.02_{\pm0.09}^{p<0.001}$ & $83.16_{\pm0.17}^{p<0.001}$ \\
    & MambaAD
    & $93.64_{\pm0.26}^{p<0.001}$ & $93.92_{\pm0.23}^{p<0.001}$ & $89.61_{\pm0.29}^{p<0.001}$ & $98.28_{\pm0.10}^{p<0.001}$ & $36.07_{\pm1.77}^{p<0.001}$ & $40.85_{\pm1.60}^{p<0.001}$ & $91.11_{\pm0.49}^{p<0.001}$ \\
    \bottomrule
  \end{tabular}%
  }
  \label{tab:statistical_robustness}
\end{table*}

\begin{table*}[t]
  \centering
  \caption{\textbf{Multi-view} multi-class UAD performance on Real-IAD~\citep{wang2024real} and MANTA~\citep{fan2025manta} (\%).}
  \resizebox{\textwidth}{!}{
    \begin{tabular}{l|cccccccc|cccccccc}
    \toprule
    \multirow{3}[2]{*}{Method} & \multicolumn{8}{c|}{\textbf{Real-IAD} (30 classes)} & \multicolumn{8}{c}{\textbf{MANTA} (38 classes)} \\
    % \cmidrule{2-9} \cmidrule{10-17}
    & Object & \multicolumn{3}{c}{Image-level} & \multicolumn{4}{c|}{Pixel-level} & Object & \multicolumn{3}{c}{Image-level} & \multicolumn{4}{c}{Pixel-level} \\
    \cmidrule(r){2-2} \cmidrule(lr){3-5} \cmidrule(lr){6-9} \cmidrule(lr){10-10} \cmidrule(lr){11-13} \cmidrule(l){14-17}
    & { AUROC} & { AUROC} & { AP} & { $F_1$-max} & { AUROC} & { AP} & { $F_1$-max} & { AUPRO} & { AUROC} & { AUROC} & { AP} & { $F_1$-max} & { AUROC} & { AP} & { $F_1$-max} & { AUPRO} \\
    \hline
    RD4AD~\citep{deng2022anomaly} & 85.7 & 83.0 & 79.6 & 74.4 & 97.3 & 25.8 & 33.5 & 90.4 & 78.1 & 80.5 & 59.7 & 62.1 & 91.9 & 28.5 &33.6 & 76.6 \\
    SimpleNet~\citep{liu2023simplenet} & 61.2 & 57.2 & 53.4 & 61.5 & 75.7 & \;\;2.8 & \;\;6.5 & 39.0 & 59.6 & 58.3 & 34.8 & 43.3 & 66.3 & \;\;5.8 & \;\;9.3 & 30.3 \\
    DeSTSeg~\citep{zhang2023destseg} & 89.0 & 82.3 & 79.2 & 73.2 & 94.6 & \underline{37.9} & \underline{41.7} & 40.6 &  66.7 &65.3 & 40.9 & 46.8 & 71.2 & \;\;9.8 & \;\;8.7 & 27.0 \\
    UniAD~\citep{you2022unified} & 87.3 & 83.0 & 80.9 & 74.3 & 97.3 & 21.1 & 29.2 & 86.7 & 78.2 & 80.5 & 59.7 & 62.5 & 89.8 & 22.6 & 29.5 & 76.3 \\
    ViTAD~\citep{zhang2023exploring} & 88.8 & 82.7 & 80.2 & 73.7 & 97.2 & 24.3 & 32.3 & 84.8 & \underline{88.9} & \underline{88.1} & \underline{75.8} & \underline{71.6} & \underline{94.9} & \underline{40.1} & \underline{43.9} & 80.6 \\
    MambaAD~\citep{he2024mambaad} & 89.9 & 86.3 & 84.6 & 77.0 & \underline{98.5} & 33.0 & 38.7 & 90.5 & 86.3 & 86.1 & 73.1 & 69.6 & 94.1 & 37.0 & 41.4 & \underline{80.7} \\
    MVAD~\citep{he2024learning} & \underline{90.2} & \underline{86.6} & \underline{84.8} & \underline{77.2} & 97.9 & 30.3 & 36.8 & \underline{91.2} & 84.4 & 84.6 & 69.4 & 67.3 & 93.5 & 32.5 & 38.2 & 80.3 \\
    \rowcolor{ours!50} Dinomaly-B~\citep{guo2025dinomaly} & 92.2 & 89.3 & 86.8 & 80.2 & 98.8 & 42.8 & 47.1 & 93.9 & 92.1 & 91.1 & 82.1 & 77.0 & 95.1 & 44.8 & 48.5 & 84.1 \\
    \rowcolor{ours} \textbf{Dinomaly2-B} & 94.3 & 91.6 & 89.8 & 82.6 & \textbf{99.2} & 47.8 & 51.1 & 96.0 & 93.2 & 92.2 & 83.7 & 79.0 & 96.1 & 51.6 & 53.1 & 88.3 \\
    \rowcolor{ours} \textbf{Dinomaly2-L} & \textbf{94.9} & \textbf{92.1} & \textbf{90.2} & \textbf{83.4} & \textbf{99.2} & \textbf{48.4} & \textbf{51.8} & \textbf{96.2} & \textbf{94.6} & \textbf{93.5} & \textbf{86.0} & \textbf{81.0} & \textbf{96.7} & \textbf{54.1} & \textbf{55.5} & \textbf{89.3} \\
    \bottomrule
    \end{tabular}
  }
  \label{tab:realiad_menta}
\end{table*}

\begin{table*}[t]
  \centering
  \caption{\textbf{Large-scale} multi-class UAD performance on Real-IAD Variety~\citep{zhu2026real} and ADNet~\citep{ling2026adnet} (\%).}
  \resizebox{\textwidth}{!}{%
  \begin{tabular}{l|cccccccc|ccccccc}
    \toprule
    \multirow{3}{*}{Method} & \multicolumn{8}{c|}{\textbf{Real-IAD Variety} (160 categories)} & \multicolumn{7}{c}{\textbf{ADNet} (380 categories)} \\
    & Object & \multicolumn{3}{c}{Image-level} & \multicolumn{4}{c|}{Pixel-level} & \multicolumn{3}{c}{Image-level} & \multicolumn{4}{c}{Pixel-level} \\
    \cmidrule(r){2-2} \cmidrule(lr){3-5} \cmidrule(lr){6-9} \cmidrule(lr){10-12} \cmidrule(l){13-16}
    & AUROC & AUROC & AP & {$F_1$-max} & AUROC & AP & {$F_1$-max} & AUPRO & AUROC & AP & $F_1$ & AUROC & AP & { $F_1$-max} & AUPRO \\
    \midrule
    RD4AD~\citep{deng2022anomaly} & 76.7 & 72.0 & 93.6 & 93.5 & 88.8 & 22.3 & 28.1 & 68.8 & 66.9 & 83.2 & 86.4 & 87.4 & 15.7 & 21.2 & 69.3 \\
    SimpleNet~\citep{liu2023simplenet} & 70.3 & 62.1 & 91.4 & 93.1 & 78.1 & 9.7 & 14.3 & 42.1 & 52.7 & 75.9 & 84.0 & 65.0 & 3.9 & 7.5 & 32.8 \\
    UniAD~\citep{you2022unified} & 67.9 & 67.1 & 92.3 & 93.3 & 87.0 & 17.9 & 23.9 & 62.1 & 63.4 & 81.5 & 85.8 & 86.7 & 11.8 & 17.4 & 66.3 \\
    ViTAD~\citep{zhang2023exploring} & 83.6 & 78.3 & 95.4 & \underline{93.7} & \underline{90.9} & \underline{31.9} & \underline{36.3} & 70.3 & 72.4 & 86.3 & \underline{87.9} & 88.1 & 19.6 & 24.4 & 68.2 \\
    MambaAD~\citep{he2024mambaad} & \underline{85.0} & \underline{79.0} & \underline{95.6} & \underline{93.7} & 90.2 & 30.9 & 35.5 & \underline{70.6} & \underline{72.7} & \underline{87.0} & 87.8 & \underline{88.8} & \underline{20.7} & \underline{25.8} & \underline{71.3} \\
    \rowcolor{ours!50} Dinomaly-B~\citep{guo2025dinomaly} & 92.5 & 87.0 & 97.5 & 94.8 & 93.5 & 44.0 & 47.0 & 79.5 & 78.1 & 89.7 & 89.8 & 90.1 & 22.3 & 28.1 & 73.3 \\
    \rowcolor{ours} \textbf{Dinomaly2-B} & 94.2 & 89.0 & 97.9 & 95.1 & 94.0 & 50.5 & 51.5 & 81.8 & 82.0 & 91.4 & 91.0 & 93.1 & 32.9 & 37.2 & 81.4 \\
    \rowcolor{ours} \textbf{Dinomaly2-L} & \textbf{94.6} & \textbf{89.3} & \textbf{98.0} & \textbf{95.2} & \textbf{94.4} & \textbf{51.8} & \textbf{52.6} & \textbf{82.8} & \textbf{83.4} & \textbf{91.8} & \textbf{91.5} & \textbf{93.7} & \textbf{34.8} & \textbf{39.2} & \textbf{82.6} \\
    \bottomrule
  \end{tabular}%
  }
  \label{tab:large_scale}
\end{table*}

\textbf{Metrics.} Following prior works ~\citep{he2024mambaad,zhang2023exploring}, we adopt seven evaluation metrics. Image-level (I-) anomaly detection performance is measured by the area under the receiver operating characteristic curve (AUROC), Average Precision (AP), and $F_{1}$ score under optimal threshold ($F_{1}$-max).  Pixel-level (P-) anomaly localization is measured by AUROC, AP, $F_{1}$-max and the Area Under the Per-Region-Overlap (AUPRO). Results for each dataset are averaged across categories. 

Anomalous pixels typically constitute a small minority. Under this imbalance, high pixel AUROC can coexist with lower AP and $F_1$-max values due to the sensitivity of precision to false positives. AUPRO complements these metrics by measuring anomaly-region coverage across false-positive rates.

 \textbf{Implementation Details.} By default, we used ViT-S/B/L pre-trained with DINOv2-registers (DINOv2-R) ~\citep{darcet2023vision} as the encoder. For the 12-layer ViT-B and ViT-S, features from the middle eight layers serve as bottleneck inputs and reconstruction targets. For the 24-layer ViT-L, we select every other layer from layers 5 to 19: $\mathcal{M}=\{5,7,9,\ldots,19\}$. The decoder always contains eight layers. Unless otherwise specified, a single model is trained jointly for all categories in each dataset.

 StableAdamW optimizer ~\citep{wortsman2023stable} is utilized with $lr$ (learning rate)=2e-3, $\beta$=(0.9,0.999), $wd$ (weight decay)=1e-4 and $eps$=1e-10.  Under the MUAD setting, the model is trained with the batch size of 16 for $It$ iterations as in Table~\ref{tab:anomaly_datasets}. The epochs can be derived by $\frac{16 \cdot It}{\#\text{Train}}$. The learning rate increases from 0 to $2\times10^{-3}$ during the first 100 iterations and then follows a cosine decay to $2\times10^{-4}$. In the Noisy Bottleneck, the dropout rate is set to 0.2 or 0.4 based on a simple hyperparameter search. Loose constraint with 2 groups is used by default. The discarding rate controlling $\tau_k$ linearly increases from 0\% to 90\% in the first 1,000 iterations as warm-up. The input image size is $392^2$ by default, and $280^2$ for newer MANTA, Real-IAD Variety, ADNet, Uni-Medical, and ApoCell. We use DINOv2's official bicubic interpolation routine to resize the pre-trained patch positional embeddings to match the input size. The mean of the top 1\% pixels in an anomaly map is used as the image-level anomaly score. For Real-IAD with tiny anomalies, we use the top 0.1\%. The method is implemented in Python 3.8 and PyTorch 1.12.0 cuda 11.3, and experiments were run on NVIDIA GeForce RTX3090/4090 GPUs (24GB). Most results for the competing MUAD methods are taken from the ADer benchmark~\citep{zhang2024ader}.   Code will be available at: \url{https://github.com/guojiajeremy/Dinomaly}.

\subsection{Comparison Results}

\subsubsection{Conventional 2D MUAD}
We first evaluate Dinomaly2 against state-of-the-art multi-class UAD methods on conventional 2D industrial datasets. As shown in Table~\ref{tab:mvtec}, Dinomaly-B (equipped with ViT-B) already exceeds previous methods by a large margin, while Dinomaly2-B further improves most of the reported metrics. Our strongest model Dinomaly2-L (equipped with ViT-L) achieves exceptional performance across all benchmarks, with I-AUROC reaching 99.9\%, 99.3\%, 99.0\%, and 97.5\% on MVTec-AD~\citep{bergmann2019mvtec}, VisA~\citep{zou2022spot}, MPDD~\citep{jezek2021deep}, and BTAD~\citep{mishra2021btad}, respectively. Compared with the previous state of the art, these results represent I-AUROC gains of 1.3, 3.8, 6.4, and 2.7 percentage points, respectively.
The category-wise results show that Dinomaly2-L performs well on both the contaminated BTAD-02 category and the clean BTAD-01 and BTAD-03 categories (Table~\ref{tab:btad_category}). These results suggest its potential for anomaly detection with noisy training data.

% Use the template's underlying table float to keep the caption outside body scaling.
\begin{tableorg}[t]
  \centering
  \footnotesize
  \setlength{\tabcolsep}{3pt}
  \caption{Category-wise BTAD performance (\%). BTAD-02 contains unidentified anomalies in its normal training split. }
  \setlength{\tabcolsep}{2pt}
  \resizebox{\columnwidth}{!}{%
  \begin{tabular}{@{}l*{12}{c}@{}}
\toprule
 & \multicolumn{4}{c}{BTAD-01} & \multicolumn{4}{c}{BTAD-02} & \multicolumn{4}{c}{BTAD-03} \\
\cmidrule(lr){2-5}\cmidrule(lr){6-9}\cmidrule(lr){10-13}
Method & I-ROC & I-AP & P-ROC & P-AP & I-ROC & I-AP & P-ROC & P-AP & I-ROC & I-AP & P-ROC & P-AP \\
\midrule
UniAD & \underline{99.8} & \underline{99.9} & \underline{97.4} & \underline{55.7} & \underline{84.0} & \underline{97.3} & 95.1 & 43.5 & \underline{99.8} & \underline{97.8} & \underline{99.7} & \underline{57.8} \\
ViTAD & 98.3 & 99.3 & 96.5 & 55.0 & \underline{84.0} & \underline{97.3} & \textbf{96.7} & \textbf{72.9} & 99.6 & 94.3 & 99.6 & 47.0 \\
MambaAD & 97.7 & 99.2 & 97.1 & 55.3 & 81.6 & 97.0 & 96.1 & 54.2 & 99.5 & 92.3 & 99.6 & 44.2 \\
\rowcolor{ours} \textbf{Dinomaly2-L} & \textbf{99.9} & \textbf{100.0} & \textbf{97.9} & \textbf{68.7} & \textbf{92.8} & \textbf{98.9} & \textbf{96.7} & \underline{72.3} & \textbf{99.9} & \textbf{98.9} & \textbf{99.9} & \textbf{80.5} \\
\bottomrule
  \end{tabular}%
  }
  \label{tab:btad_category}
  \vspace{-5pt}
\end{tableorg}

MVTec-AD and VisA are the two most widely used industrial UAD benchmarks, where performance is approaching a plateau. Consequently, the incremental pixel-level gains of Dinomaly2-B over the already strong Dinomaly-B baseline are relatively limited on these datasets. The improvements become more pronounced on the more challenging datasets evaluated in the following sections.

To assess the reproducibility of the gains on MVTec-AD and VisA, we repeat Dinomaly2-B, Dinomaly-B, ViTAD, and MambaAD with five random seeds. As shown in Table~\ref{tab:statistical_robustness}, the small image-level improvement over Dinomaly-B on MVTec-AD remains statistically significant, whereas their pixel-level differences are not uniformly significant. Dinomaly2-B significantly outperforms both ViTAD and MambaAD across all seven metrics on MVTec-AD and VisA.

% ── MVTec LOCO ──────────────────────────────────────────────────

\begin{table}[t]
  \centering
  \tiny
  \setlength{\tabcolsep}{0.3pt}
  \caption{Multi-class UAD performance on MVTec LOCO~\citep{bergmann2022beyond} (\%). The lower block reports structural (Str.) and logical (Log.) anomalies separately for Dinomaly variants.}
  \begin{tabular*}{\columnwidth}{@{\extracolsep{\fill}}lccccccc@{}}
	    \toprule
	    \multirow{2}{*}{Method} & \multicolumn{3}{c}{Image-level} & \multicolumn{4}{c}{Pixel-level} \\
	    \cmidrule(lr){2-4} \cmidrule(lr){5-8}
	    & AUROC & AP & $F_1$-max & AUROC & AP & $F_1$-max & AUPRO \\
	    \midrule
	    RD4AD & 73.7 & 84.6 & 79.4 & 70.7 & 21.0 & 26.2 & 67.5 \\
	    SimpleNet & \underline{81.8} & \underline{89.1} & \underline{82.5} & 70.9 & 28.3 & 32.1 & 61.2 \\
	    DeSTSeg & 81.2 & 88.6 & 81.8 & 63.7 & 31.8 & 31.9 & 59.2 \\
	    UniAD & 78.7 & 87.1 & 80.9 & 74.6 & \underline{32.6} & \underline{35.8} & \textbf{70.7} \\
	    DiAD & 77.2 & 84.6 & 81.5 & 72.1 & 18.6 & 25.3 & 54.4 \\
	    ViTAD & 76.2 & 85.0 & 80.1 & 73.2 & 27.9 & 30.6 & 62.0 \\
	    MambaAD & 78.2 & 87.1 & 81.8 & \textbf{77.5} & 28.0 & 32.9 & 68.6 \\
	    \rowcolor{ours!50} Dinomaly  & 82.0 & 89.1 & 82.8 & 75.7 & 34.0 & 36.4 & 63.0 \\
	    \rowcolor{ours} \textbf{Dinomaly2-B} & 84.4 & 91.0 & \textbf{84.0} & 77.0 & 37.8 & 42.4 & 68.5 \\
	    \rowcolor{ours} \textbf{Dinomaly2-L} & \textbf{85.4} & \textbf{91.8} & 83.9 & \textbf{77.5} & \textbf{40.9} & \textbf{45.5} & \underline{70.0} \\
	    \midrule
	    \multicolumn{8}{l}{\textit{Structural/logical breakdown of Dinomaly variants}} \\
	    \rowcolor{ours!50} Dinomaly-B (Str.) & 92.6 & 91.3 & 86.4 & 96.6 & 54.6 & 57.9 & 92.4 \\
	    \rowcolor{ours!50} Dinomaly-B (Log.) & 74.4 & 76.1 & 72.6 & 75.0 & 35.4 & 38.6 & 51.6 \\
	    \rowcolor{ours} Dinomaly2-B (Str.) & 95.6 & 95.0 & 90.9 & 97.3 & 54.9 & 59.0 & 93.8 \\
	    \rowcolor{ours} Dinomaly2-B (Log.) & 76.2 & 78.3 & 74.6 & 76.2 & 40.1 & 43.4 & 59.3 \\
	    \rowcolor{ours} Dinomaly2-L (Str.) & 97.1 & 96.9 & 93.1 & 97.9 & 57.7 & 61.7 & 94.7 \\
	    \rowcolor{ours} Dinomaly2-L (Log.) & 76.6 & 78.3 & 74.1 & 76.7 & 43.2 & 46.1 & 61.1 \\
	    \bottomrule
	  \end{tabular*}
  \label{tab:loco}
\end{table}

\subsubsection{Multi-View MUAD} 
Real-world industrial inspection often requires examining objects from multiple viewpoints to ensure comprehensive defect detection. On the Real-IAD~\citep{wang2024real} and MANTA~\citep{fan2025manta}, Dinomaly2 demonstrates superior performance across all evaluation levels. As presented in Table~\ref{tab:realiad_menta}, our method achieves 94.9\% object-level AUROC on Real-IAD and 94.6\% object-level AUROC on MANTA. These results substantially outperform previous methods, including the MVAD ~\citep{he2024learning} with multi-view interaction operation designed specifically for multi-view inspection.

\subsubsection{Large-Scale MUAD}
Dinomaly2 remains effective in large-scale MUAD with hundreds of categories. On Real-IAD Variety~\citep{zhu2026real} (160 categories), Dinomaly2-B improves I-AUROC and P-AP by 10.0 and 18.6 percentage points, respectively, over the strongest non-Dinomaly result for each metric. The corresponding gains on ADNet~\citep{ling2026adnet} (380 categories) are 9.3 and 12.2 points (Table~\ref{tab:large_scale}). It also improves every reported image- and pixel-level metric over Dinomaly-B on both datasets. Dinomaly2-L further improves performance, reaching 89.3\% and 83.4\% I-AUROC, respectively.

\subsubsection{Logical MUAD}
Table~\ref{tab:loco} compares unified models on MVTec LOCO, which contains both structural and logical anomalies. Among the compared unified UAD methods, Dinomaly2 achieves the best or tied-best results on six of the seven overall metrics. Dinomaly2-B improves over Dinomaly-B by 2.4 and 5.5 percentage points in I-AUROC and P-AUPRO, respectively. The structural-logical breakdown nevertheless reveals that detecting logical anomalies remains challenging in the unsupervised setting. These results suggest that better modeling of component-level relationships could further improve logical anomaly detection.

% ── RGB-3D ──────────────────────────────────────────────────

\begin{table}[t]
  \centering
 \tiny 
  \setlength{\tabcolsep}{3pt}
  \caption{{Multi-class} UAD performance for \textbf{RGB-3D} inspection on {MVTec3D}~\citep{bergmann2021mvtec3d} with 10 classes (\%). Results are reported for both RGB-only and RGB–3D Dinomaly2. \dag~: methods specifically  designed for RGB+3D UAD.}
      \begin{tabular}{llccccccc}
    \toprule
  &   Method & I-ROC & P-ROC & P-PRO \\   

\hline
\multirow{7}[0]{*}{\rotatebox{90}{{RGB}}}  &   RD4AD~\citep{deng2022anomaly} & 79.3& 98.3& 93.5\\
  &   DeSTSeg~\citep{zhang2023destseg} & 82.5& 82.1& 65.1\\
  &   UniAD~\citep{you2022unified} & 77.0& 96.8 & 89.4\\
  &   ViTAD~\citep{zhang2023exploring} & 79.7 & 98.0 & 91.3 \\
  &   MambaAD~\citep{he2024mambaad} & \underline{85.8} & \underline{98.6} & \underline{94.1}\\
  &  \cellcolor{ours}\textbf{Dinomaly2-B} & \cellcolor{ours}{93.9} & \cellcolor{ours}\textbf{99.2}& \cellcolor{ours}\textbf{97.4} \\
   &  \cellcolor{ours}\textbf{Dinomaly2-L} & \cellcolor{ours}\textbf{95.0}& \cellcolor{ours}{99.1}& \cellcolor{ours}{97.0}\\
\hline

\multirow{7}[0]{*}{\rotatebox{90}{{RGB+3D}}} &   BTF\dag~\citep{horwitz2023back} & 68.2& 96.8& 90.4\\
 &   M3DM\dag~\citep{wang2023multimodal} & 72.4& 96.0& 87.2\\
 &   Shape-G\dag~\citep{chu2023shape} & 89.8& 97.5& 92.5\\
 &   CPMF\dag~\citep{cao2024complementary} & 91.8& 96.8& 90.4\\
 &   GLFM\dag~\citep{cheng2025boosting} & \underline{94.4} & \underline{97.9} & \underline{93.1}\\
 &  \cellcolor{ours}\textbf{Dinomaly2-B} & \cellcolor{ours}97.0& \cellcolor{ours}\textbf{99.6}& \cellcolor{ours}\textbf{98.6}\\
 &  \cellcolor{ours}\textbf{Dinomaly2-L} & \cellcolor{ours}\textbf{97.4}& \cellcolor{ours}\textbf{99.6}& \cellcolor{ours}\textbf{98.6}\\

    \bottomrule
    \end{tabular}%
  \label{tab:mvtec3d}%
\end{table}%

% ── RGB-IR ──────────────────────────────────────────────────

\begin{table}[t]
  \centering
 \tiny 
  \setlength{\tabcolsep}{3pt}
  \caption{{Multi-class} UAD performance for \textbf{RGB-IR} inspection on MulSen-AD~\citep{li2025multi} with 15 classes (\%). \dag~: methods specifically designed for multi-sensor UAD. \textit{MC}: multi-class unified. O-ROC: object-level AUROC. Pixel-level metrics are calculated respectively on RGB/Infra-red image. }
      \begin{tabular*}{\columnwidth}{@{\extracolsep{\fill}} lclcccccc @{}}
    \toprule
 & \textit{MC} &    Method & O-ROC & P-ROC & P-AP \\   

\hline
\multirow{3}[0]{*}{\rotatebox{90}{{RGB}}}  &  \textcolor{red}{\ding{55}} & PatchCore~\citep{roth2022towards} & 83.7& - & - \\
  & \textcolor{red}{\ding{55}} &  RD++~\citep{deng2022anomaly} & 86.2 & - & - \\
  & \textcolor{red}{\ding{55}} &  SimpleNet~\citep{liu2023simplenet} & 85.3 & - & -\\
\hline

\multirow{6}[0]{*}{\rotatebox{90}{{RGB+IR}}} & \textcolor{red}{\ding{55}} &  TripleAD\dag~\citep{li2025multi} & 94.8 & 98.2/97.2& 34.3/32.2\\
    & \checkmark &  RD4AD~\citep{deng2022anomaly} & 85.0& \underline{98.5/98.3} & 27.4/28.2\\
    & \checkmark &  ViTAD~\citep{zhang2023exploring} & 90.5& 97.6/96.3& 22.6/21.2\\
    & \checkmark &  TripleAD\dag~\citep{li2025multi} & \underline{94.2} & 98.3/97.9& \underline{37.2/32.6} \\
 & \checkmark &  \cellcolor{ours} \textbf{Dinomaly2-B} & \cellcolor{ours}\textbf{97.6}& \cellcolor{ours}\textbf{99.2}/{98.7}& \cellcolor{ours}\textbf{39.8}/{41.1}\\

  & \checkmark &  \cellcolor{ours} \textbf{Dinomaly2-L} & \cellcolor{ours}{97.0}& \cellcolor{ours}\textbf{99.2}/\textbf{98.8}& \cellcolor{ours}{39.3}/\textbf{41.8}\\
  
    \bottomrule
    \end{tabular*}%
    
  \label{tab:mulsen}%
\vspace{-5pt}
\end{table}%

% ── Single-class ──────────────────────────────────────────────

\begin{table}[t]
  \centering
 \tiny 
  \setlength{\tabcolsep}{3pt}
  \caption{\textbf{Single-class} UAD performance (\%).Multi-class Dinomaly2 results are also presented for comparison. \textit{MC}: multi-class unified. }
      \begin{tabular*}{\columnwidth}{@{\extracolsep{\fill}} lclcccccc @{}}
    \toprule
 & MC &  Method & I-ROC & P-ROC & P-PRO \\   

\hline
\multirow{5}[0]{*}{\rotatebox{90}{{MVTec}}}  & \textcolor{red}{\ding{55}} & RD4AD~\citep{deng2022anomaly} & 98.5& 97.8& 93.9\\
& \textcolor{red}{\ding{55}} &   PatchCore~\citep{roth2022towards} & 99.1 & 98.1& 93.5\\
& \textcolor{red}{\ding{55}} &   SimpleNet~\citep{hyun2024reconpatch} & 99.6& 98.1& 90.0\\
& \textcolor{red}{\ding{55}} &  \cellcolor{ours!50}\textbf{Dinomaly2-B} & \cellcolor{ours!50}\textbf{99.8} & \cellcolor{ours!50}\textbf{98.4}& \cellcolor{ours!50}\textbf{95.3} \\
& \checkmark &  \cellcolor{ours}Dinomaly2-B & \cellcolor{ours}\textit{99.8} & \cellcolor{ours}\textit{98.4}& \cellcolor{ours}\textit{95.1} \\
\hline

\multirow{5}[0]{*}{\rotatebox{90}{{VisA}}} & \textcolor{red}{\ding{55}} &   RD4AD~\citep{deng2022anomaly} &  96.0& 90.1& 70.9\\
& \textcolor{red}{\ding{55}} &   GLASS~\citep{chen2024unified} & 98.8 & 98.8& 92.8\\
& \textcolor{red}{\ding{55}} &   DiffusionAD~\citep{zhang2025diffusionad} & 98.8& 98.9& 96.0\\
& \textcolor{red}{\ding{55}} &  \cellcolor{ours!50}\textbf{Dinomaly2-B} & \cellcolor{ours!50}\textbf{99.2} & \cellcolor{ours!50}\textbf{99.2}& \cellcolor{ours!50}\textbf{95.7} \\
&  \checkmark &  \cellcolor{ours}Dinomaly2-B & \cellcolor{ours}\textit{99.2} & \cellcolor{ours}\textit{99.1}& \cellcolor{ours}\textit{95.5} \\
% &  \checkmark &  \cellcolor{ours}Dinomaly2-B & \cellcolor{ours}\textit{93.4} & \cellcolor{ours}\textit{99.3}& \cellcolor{ours}\textit{96.5} \\
    \bottomrule
    \end{tabular*}%
    
  \label{tab:single}%
\end{table}%

% ── Few-shot ──────────────────────────────────────────────────

\begin{table}[t]
  \centering
 \tiny 
  \setlength{\tabcolsep}{3pt}
  \caption{ \textbf{Few-shot} multi-class  UAD performance (\%). The multi-class unified model is built using K-shot per class. Results of Dinomaly2 are the average over five experiments with random selected normal shots. \dag~denotes method specifically designed for few-shot UAD.}
      \begin{tabular*}{\columnwidth}{@{\extracolsep{\fill}} cclcccccc @{}}
    \toprule
  & Shot &  Method & I-ROC & P-ROC & P-PRO \\   

\hline
\multirow{8}[0]{*}{\rotatebox{90}{\textbf{MVTec-AD}}} & \multirow{6}[0]{*}{4}  &   PatchCore~\citep{roth2022towards} & 74.9 &  92.6 & 80.8\\
&  &   WinCLIP\dag~\citep{jeong2023winclip} &  94.0 & 92.9 & 84.4\\
&  &   PromptAD\dag~\citep{li2024promptad} &   90.6 &  92.4 & 84.6\\
&  &   InCTRL\dag~\citep{zhu2024toward} &  94.5 &  - & -\\
&  &   IIPAD\dag~\citep{iipad}\dag &  \underline{96.1} & \underline{97.0} & \underline{91.2}\\

&  &  \cellcolor{ours}\textbf{Dinomaly2-B} & \cellcolor{ours}\textbf{98.1} &  \cellcolor{ours}\textbf{97.4}& \cellcolor{ours}\textbf{93.5} \\ \cmidrule(lr){2-6}
&  8&   \textbf{Dinomaly2-B} &  98.7 & 97.5 & 93.6\\
&  16& \textbf{Dinomaly2-B} &  99.0 & 97.6 & 93.7\\
&  Full& MambaAD~\citep{he2024mambaad} &  98.6 & 97.7 & 93.1\\

\midrule

\multirow{8}[0]{*}{\rotatebox{90}{\textbf{VisA}}}  &  \multirow{6}[0]{*}{4} &  PatchCore~\citep{roth2022towards} & 62.6 &  85.4 & 70.6\\
&  &   WinCLIP\dag~\citep{jeong2023winclip} &  86.1 & 95.2 & 82.1\\
&  &   PromptAD\dag~\citep{li2024promptad} &  88.8 &  97.2 & 84.7\\
&  &   InCTRL\dag~\citep{zhu2024toward} &  \underline{90.2} &  - & -\\
&  &   IIPAD\dag~\citep{iipad} &  88.3 & \underline{97.4} & \underline{88.3}\\

&  &  \cellcolor{ours}\textbf{Dinomaly2-B} & \cellcolor{ours}\textbf{96.7} &  \cellcolor{ours}\textbf{98.2}& \cellcolor{ours}\textbf{94.4} \\ \cmidrule(lr){2-6}
& 8 &  \textbf{Dinomaly2-B} & 97.4 &  98.4&  94.8 \\
& 16 &  \textbf{Dinomaly2-B} & 98.0 &  98.4& 94.8 \\
&  Full& ReContrast~\citep{guo2023recontrast} &  95.5 & 98.5 & 91.9\\

\midrule
\multirow{8}[0]{*}{\rotatebox{90}{\textbf{MVTec3D}}} & \multirow{5}[0]{*}{4}  &   BTF~\citep{horwitz2023back} & 64.3 &  97.6  & 90.4\\
&  &   Shape-G~\citep{chu2023shape} &  69.8 & \underline{97.9} & \underline{91.8} \\
&  &   AST~\citep{rudolph2023asymmetric} &   53.5 &  91.4 & 73.9\\
&  &   CLIP3D-AD\dag~\citep{zuo2024clip3d} &  \underline{74.0} &  97.0 & 90.2\\
&  &  \cellcolor{ours}\textbf{Dinomaly2-B} & \cellcolor{ours}\textbf{90.3} &  \cellcolor{ours}\textbf{99.3}& \cellcolor{ours}\textbf{97.6} \\ \cmidrule(lr){2-6}
&  8&   \textbf{Dinomaly2-B} &  93.4 & 99.5 & 98.1\\
&  16& \textbf{Dinomaly2-B} &  94.2 & 99.5 & 98.3\\
&  Full& GLFM~\citep{cheng2025boosting} &  94.4 & 97.9 & 93.1\\

    \bottomrule
    \end{tabular*}%
    
  \label{tab:fewshot}%
\end{table}%

\subsubsection{RGB-3D MUAD}
Extending beyond 2D imagery, we evaluate Dinomaly2 on MVTec3D~\citep{bergmann2021mvtec3d}, which combines RGB images with 3D point cloud data for industrial inspection. As shown in Table~\ref{tab:mvtec3d}, our RGB-only model already achieves 95.0\% I-AUROC, which not only surpasses previous RGB-based methods by a significant margin, but also reaches competitive performance compared to SOTA methods that leverage RGB+3D input (94.4\%).

More importantly, our simple RGB-3D fusion strategy, \textit{i.e.}, element-wise averaging of features from independently processed RGB images and depth maps, achieves unprecedented results. Dinomaly2-L attains 97.4\% I-AUROC and 98.6\% AUPRO, outperforming previous multi-modal-specific multi-class SOTA GLFM (94.4\%, 93.1\%) by a large margin. This remarkable performance validates our hypothesis that the minimalist philosophy naturally scales across data modalities without requiring specialized architectural modifications.

\subsubsection{RGB-IR MUAD}
For industrial scenarios requiring thermal information, we evaluate Dinomaly2 on MulSen-AD. As shown in Table~\ref{tab:mulsen}, our straightforward approach achieves exceptional performance. Dinomaly2-B attains 97.6\% object-level AUROC, substantially outperforming the specialized multi-sensor method TripleAD (94.2\%).

%  ───────────── Inference-unified ─────────────

\begin{figure*}[t]
	\centering
	
	\begin{minipage}[c]{0.4\textwidth}
		\centering
		\includegraphics[width=\linewidth]{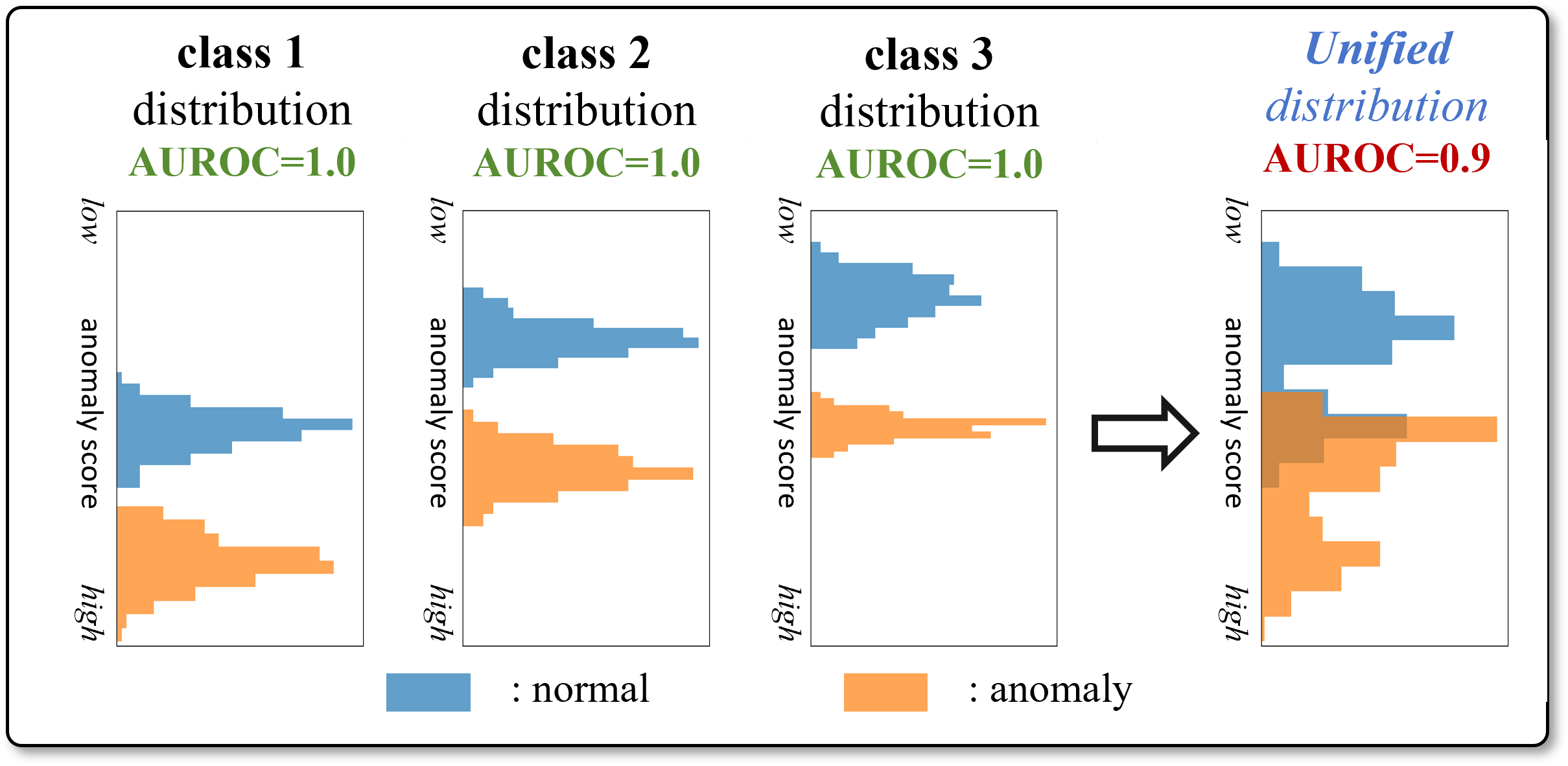}
		\setlength{\abovecaptionskip}{2pt} 
		\caption{The challenge of inference-unified multi-class UAD.}
		\label{fig:inferuni}
	\end{minipage}
	\hfill
	\begin{minipage}[c]{0.58\textwidth}
		\centering
		\makeatletter
		\def\@captype{table}
		\makeatother
		\caption{\textbf{Inference-unified} multi-class UAD performance (\%). The unified model is evaluated with all images in a dataset mixed together. Metrics represent performance across the entire mixed dataset rather than category-wise averages.}
		\label{tab:inferuni}
		
		\resizebox{\linewidth}{!}{
			\begin{tabular}{lcccccc}
				\toprule
				Method & \multicolumn{3}{c}{\textbf{MVTec-AD (Mixed)}} & \multicolumn{3}{c}{\textbf{VisA (Mixed)}} \\   
				\cmidrule(lr){2-4} \cmidrule(lr){5-7}
				& I-ROC & P-ROC & P-AP & I-ROC & P-ROC & P-AP \\
				\midrule
				RD4AD~\citep{deng2022anomaly} & 85.0& 94.5& 49.0 & 83.9& 94.9& 33.5\\
				DRAEM~\citep{zavrtanik2021draem} & 82.3 & 81.8 & 41.1 & 70.0 & 82.7 & 27.0\\
				PatchCore~\citep{roth2022towards} & 89.0 & 84.4 & 25.6 & 84.1 & 94.0 & 33.1\\
				UniAD~\citep{hyun2024reconpatch} & 91.1& 96.4& 51.1 & 89.0& 97.9& 34.7\\
				ReContrast~\citep{guo2023recontrast} & \underline{90.7} & \underline{96.8} & \underline{56.3} & \underline{93.2} & \underline{97.1}& \underline{42.1}\\  
				ViTAD~\citep{zhang2023exploring} & 88.7& 95.3& 48.8 & 84.7&  95.3& 39.0\\
				\cellcolor{ours}\textbf{Dinomaly2-B} & \cellcolor{ours}\textbf{98.9} & \cellcolor{ours}\textbf{97.3}& \cellcolor{ours}\textbf{62.4} & \cellcolor{ours}\textbf{97.8} & \cellcolor{ours}\textbf{98.7}& \cellcolor{ours}\textbf{48.6} \\
				\bottomrule
			\end{tabular}%
		}
	\end{minipage}
	
\end{figure*}

\begin{figure*}[t]
	\centering
	\includegraphics[width=\textwidth]{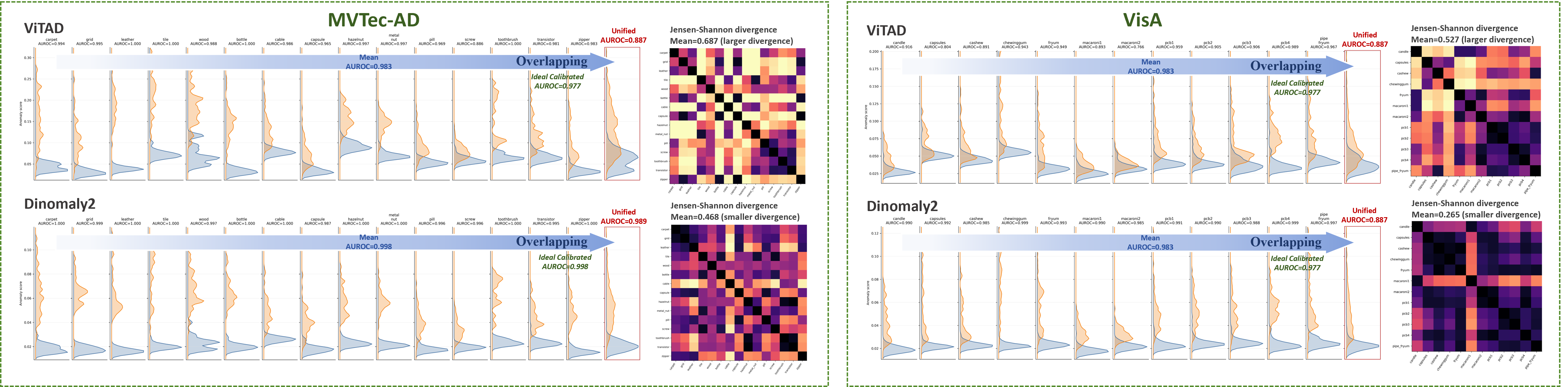}
	\caption{Inter-class score analysis under inference-unified MUAD. Normal and anomalous image-score distributions are shown for individual categories and for all categories unified. Mean I-AUROC denotes the category-wise average, while Ideal Calibrated AUROC is obtained by category-wise z-score standardization using normal-score distributions before pooling. The heatmaps report pairwise JSD between category-wise normal-score distributions; lower values indicate better inter-class alignment. ViTAD results are reproduced.}
	\label{fig:inferuni_analysis}
	\vspace{-5pt}
\end{figure*}

\subsubsection{Single-Class UAD}
To assess whether a multi-class model can match specialized single-class methods, we compare our multi-class-trained Dinomaly2 with methods trained separately for each category. As reported in Table~\ref{tab:single}, our multi-class-trained model matches or surpasses dedicated single-class methods. This result demonstrates that Dinomaly2 successfully bridges the long-standing performance gap between multi-class and single-class UAD methods, supporting the practical viability of unified models for real-world deployment.

\subsubsection{Few-Shot MUAD}
In resource-constrained scenarios where only limited normal samples are available, Dinomaly2 demonstrates exceptional few-shot learning capabilities. Using only 4 normal samples per class, our method achieves 98.1\% I-AUROC on MVTec-AD and 96.7\% on VisA, as shown in Table~\ref{tab:fewshot}. These results even outperform the method that was specially designed for multi-class few-shot anomaly detection (IIPAD~\citep{iipad}: 96.1\%, 88.3\%). Moreover, our approach seamlessly integrates few-shot UAD with RGB-3D data. On the MVTec3D dataset, our method achieves an I-AUROC of 90.3\% using only 4 samples per class, substantially outperforming the current SOTA~\citep{zuo2024clip3d} (74.0\%).

%  ───────────── Compare with other extensions ─────────────

\begin{table*}[t]
	\centering
	\tiny
	\begin{minipage}[c]{0.35\linewidth}
		\caption{Comparison with new methods integrated on Dinomaly~\citep{guo2025dinomaly}. Results are presented as I-AUROC/AUPRO.}
		\label{tab:dbased}
	\end{minipage}
	\hfill
	\begin{minipage}[c]{0.64\linewidth}
		\begin{tabular}{lccc}
			\toprule
			Method & \textbf{MVTec-AD} & \textbf{VisA} & \textbf{Real-IAD} \\
			\hline
			\rowcolor{ours!50} Dinomaly-B~\citep{guo2025dinomaly}             & 99.6/94.8          & 98.7/94.5          & 89.3/93.9 \\
			\rowcolor{ours!50} ~~+In-Normal Prototype~\citep{luo2025exploring}                & 99.7/94.9          & 98.9/94.4          & 90.5/95.0 \\
			\rowcolor{ours!50} ~~+CostFilter~\citep{zhang2025costfilter}      & 99.7/94.8          & 98.7/94.7          & -         \\
			\rowcolor{ours!50} ~~+Inter-View Attn.~\citep{zhou2025multi}  &  -                & -                  & 89.7/94.4 \\
			\rowcolor{ours!50} ~~+Entropy-Increased LA.~\citep{liu2026entropy}  & 99.8/95.4                   & 98.9/95.4                   & 89.8/94.3 \\
			\rowcolor{ours}    \textbf{Dinomaly2-B}                          & \textbf{99.8/95.1} & \textbf{99.2/95.5} & \textbf{91.1/96.0} \\ \hline
			\rowcolor{ours}    \textbf{Dinomaly2-L}                          & \textbf{99.9/95.1} & \textbf{99.3/95.4} & \textbf{92.1/96.2} \\
			\bottomrule
		\end{tabular}
	\end{minipage}
	\vspace{-5pt}
\end{table*}

%  Uni-Medical / MIAD / Drone-Anomaly / ApoCell ─────────────
\begin{table*}[t]
	\centering
	\footnotesize
	\caption{Multi-class UAD performance on Uni-Medical~\citep{zhang2024ader}, ApoCell~\citep{guan2025cell}, MIAD~\citep{bao2023miad}, and Drone-Anomaly~\citep{jin2022anomaly} (\%).}
	\begin{minipage}[t]{0.48\linewidth}
		\centering
		\begin{tabular}{lccc}
			\toprule
			Method & I-ROC & P-ROC & P-PRO \\
			\midrule
			\multicolumn{4}{c}{\textbf{Uni-Medical}} \\
			\hline
			RD4AD~\citep{deng2022anomaly}         & 76.4          & 96.4          & \underline{86.5} \\
			DeSTSeg~\citep{zhang2023destseg}      & 82.0          & 83.2          & 66.1 \\
			UniAD~\citep{you2022unified}          & 80.4          & 96.5          & 85.8 \\
			DiAD~\citep{he2024diad}              & \underline{82.9} & 96.0       & 85.4 \\
			ViTAD~\citep{zhang2023exploring}      & 81.5          & \underline{97.0} & \underline{86.5} \\
			\rowcolor{ours}\textbf{Dinomaly2-B}  & \textbf{86.7} & \textbf{97.9} & \textbf{90.0} \\
			\midrule
			\multicolumn{4}{c}{\textbf{MIAD}} \\
			\hline
			RD4AD~\citep{deng2022anomaly}         & 64.2          & 84.1          & \underline{69.3} \\
			DeSTSeg~\citep{zhang2023destseg}      & 68.7          & 72.6          & 16.7 \\
			UniAD~\citep{you2022unified}          & 61.2          & \underline{91.1} & 69.0 \\
			ReContrast~\citep{guo2023recontrast}  & 68.9          & 86.7          & 72.9 \\
			ViTAD~\citep{zhang2023exploring}      & \underline{71.5} & 80.9       & 60.4 \\
			\rowcolor{ours}\textbf{Dinomaly2-B}  & \textbf{84.5} & \textbf{92.1} & \textbf{81.5} \\
			\bottomrule
		\end{tabular}
	\end{minipage}
	\hfill
	\begin{minipage}[t]{0.48\linewidth}
		\centering
		\begin{tabular}{lccc}
			\toprule
			Method & I-ROC & \;\;I-AP & \;\;I-F1 \\
			\midrule
			\multicolumn{4}{c}{\textbf{Drone-Anomaly}} \\
			\hline
			ANDT\dag~\citep{jin2022anomaly}       & 68.6          & -             & - \\
			MADViT\dag~\citep{balwant2025madvit}  & 74.2          & -             & - \\
			RD4AD~\citep{deng2022anomaly}         & 68.3          & 36.9          & 48.3 \\
			UniAD~\citep{you2022unified}          & 70.4          & 44.5          & 53.8 \\
			ViTAD~\citep{zhang2023exploring}      & \underline{77.0} & \underline{56.4} & \underline{59.4} \\
			\rowcolor{ours}\textbf{Dinomaly2-B}  & \textbf{84.8} & \textbf{67.0} & \textbf{73.6} \\
			\midrule
			\multicolumn{4}{c}{\textbf{ApoCell}} \\
			\hline
			RD4AD~\citep{deng2022anomaly}         & 80.4          & \underline{70.3} & 68.7 \\
			DeSTSeg~\citep{zhang2023destseg}      & 48.1          & 36.3          & 56.7 \\
			UniAD~\citep{you2022unified}          & 47.3          & 35.2          & 56.1 \\
			ReContrast~\citep{guo2023recontrast}  & \underline{81.6} & \underline{70.3} & \underline{71.1} \\
			ViTAD~\citep{zhang2023exploring}      & 75.1          & 63.3          & 65.1 \\
			\rowcolor{ours}\textbf{Dinomaly2-B}  & \textbf{83.2} & \textbf{71.3} & \textbf{73.8} \\
			\bottomrule
		\end{tabular}
	\end{minipage}
	\label{tab:medical}
	\vspace{-5pt}
\end{table*}

\subsubsection{Inference-Unified MUAD}
Existing multi-class evaluation protocols often employ class-separate inference and evaluation. Here, we introduce the inference-unified setting where models must detect anomalies across mixed categories using the same threshold as if there is only one scene. A key challenge in this setting is the mismatch between anomaly-score distributions across object categories. Although the normal and anomalous samples are well separated within each category, they become inseparable in the unified score distribution, as shown in Fig.~\ref{fig:inferuni}.

As shown in Table~\ref{tab:inferuni}, Dinomaly2-B achieves 98.9\% I-AUROC on MVTec-AD and 97.8\% on VisA under this challenging setting, significantly outperforming previous methods.
This evaluation better reflects real-world deployment scenarios where models encounter mixed object categories without prior knowledge of class identity.

Figure~\ref{fig:inferuni_analysis} compares category-wise and unified image-score distributions. ViTAD exhibits larger category-dependent shifts in normal scores, causing greater normal--anomaly overlap after pooling, whereas Dinomaly2 remains better aligned. We quantify this mismatch by the mean off-diagonal Jensen--Shannon divergence (JSD) between category-wise normal-score distributions. JSD ranges from 0 to 1, where 0 indicates identical distributions and 1 indicates maximal difference. The JSDs of ViTAD/Dinomaly2 are 0.687/0.468 on MVTec-AD and 0.527/0.265 on VisA. We also report \textit{Ideal Calibrated AUROC}, obtained by category-wise z-score calibration, as an upper-bound reference for unified I-AUROC. Dinomaly2 is much closer to this reference than ViTAD, further indicating better inter-class score alignment.

\subsubsection{Comparison with Other Extensions}
The simplicity of the Dinomaly framework has made it easy to reuse and extend. As summarized in Table~\ref{tab:dbased}, recent works either build directly on Dinomaly or integrate their extensions into its framework, including INP-Former~\citep{luo2025exploring}, CostFilter-AD~\citep{zhang2025costfilter}, Multi-View Attention~\citep{zhou2025multi}, and EILA-AD~\citep{liu2026entropy}. These developments provide independent evidence of the extensibility of the proposed minimalist design. Dinomaly2 achieves strong overall performance relative to these extensions while retaining its minimalist design.

\begin{figure*}[!t]
	\setlength{\abovecaptionskip}{0mm}
	\centerline{\includegraphics[width=0.99\textwidth]{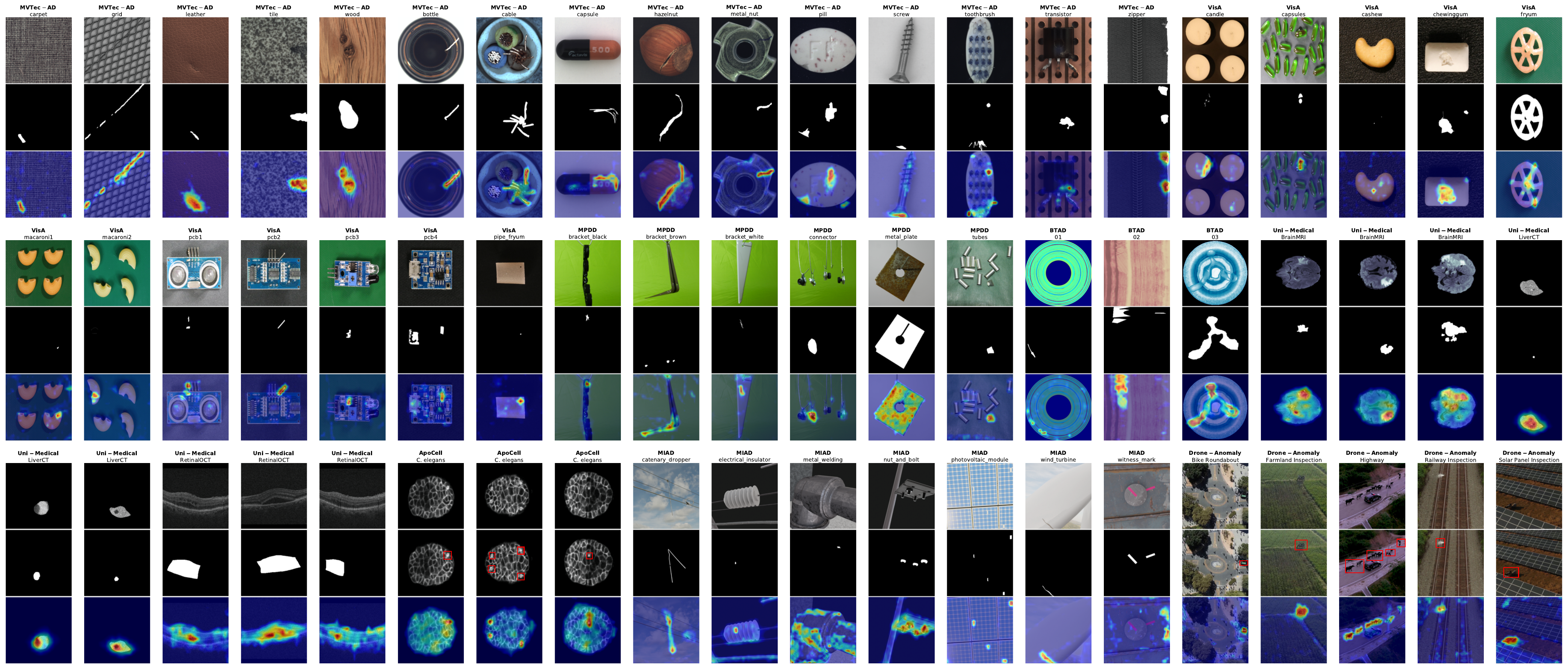}}
	\caption{Qualitative anomaly localization results across diverse domains. Anomaly maps are mean-max normalized to [0,1] range with red indicating high anomaly likelihood. All samples are randomly selected.}
	\label{fig:visualize}
	\vspace{-5pt}
\end{figure*}

\subsubsection{Beyond Industrial Domain}
We further evaluate Dinomaly2 across diverse application domains, as shown in Table~\ref{tab:medical}. On the medical imaging dataset Uni-Medical~\citep{zhang2024ader}, our method achieves SOTA performance for detecting pathological conditions in brain CT, liver CT, and retinal OCT scans. In microscopic images for biological research~\citep{guan2025cell}, Dinomaly2 attains 83.2\% I-AUROC in identifying apoptotic cells in \textit{C. elegans} embryos. In outdoor maintenance scenarios on MIAD~\citep{bao2023miad} with uncontrolled environmental factors, our method reaches 84.5\% I-AUROC.  For aerial surveillance on the Drone-Anomaly dataset ~\citep{jin2022anomaly}, Dinomaly2 successfully identifies abnormal events in UAV imagery across 7 distinct scenes.
These cross-domain results demonstrate that the minimalist design philosophy and foundational feature representations enable effective anomaly detection across vastly different data distributions and application contexts, positioning Dinomaly2 as a unified solution for real-world anomaly detection applications.

\subsubsection{Qualitative Visualization}
Figure~\ref{fig:visualize} shows the anomaly maps predicted by Dinomaly2. Anomaly maps are normalized to [0, 1] on a per-image basis using mean-max normalization.

\subsection{Ablation Study}
\subsubsection{Overall Component Analysis} 
We systematically evaluate the effectiveness of each proposed component through comprehensive ablation studies on MVTec-AD and VisA datasets. As shown in Table~\ref{tab:ablation}, our baseline achieves competitive performance of 98.73\% I-AUROC on MVTec-AD and 96.23\% on VisA, demonstrating the strong foundation provided by our simple architecture.
Each component contributes meaningfully to the overall performance. The Noisy Bottleneck (NB) provides a foundational gain, validating our hypothesis that simple Dropout-based noise injection effectively prevents over-generalization in multi-class settings. Context-Aware Recentering (CAR) shows strong improvements on VisA (+1.19\% I-AUROC), where the multi-class confusion problem is more pronounced.
Unfocused Linear Attention (LA) and Loose Constraint (LC) demonstrate synergistic effects with other components. While LA and LC individually provide modest gains, their combination with NB and other components yields substantial improvements. Loose Loss (LL) consistently improves performance across both datasets, which validates our selective optimization strategy that reduces gradient flow for well-reconstructed regions.
The full combination of all components achieves the best performance: 99.74\%/99.82\%/99.30\% on MVTec-AD and 98.89\%/99.02\%/96.47\% on VisA with only 10,000 training iterations, demonstrating that Dinomaly2 effectively addresses the challenges in multi-class anomaly detection through simple yet effective designs.

\begin{table*}[t]
	\centering
	\caption{Systematic ablation study of proposed components on MVTec-AD and VisA (\%). 
		NB: Noisy Bottleneck. LA: Unfocused Linear Attention. LC: Loose Constraint (2 groups). 
		LL: Loose Loss. CAR: Context-Aware Recentering. ViT-B is used as the backbone. 
		Training is reduced to 10,000 iterations for efficiency. 
		Image-level metrics are AUROC, AP, and $F_1$-max, while pixel-level metrics are 
		AUROC, AP, $F_1$-max, and AUPRO.}
	
	\resizebox{\textwidth}{!}{
	\begin{tabular}{@{}ccccc|ccc|cccc|ccc|cccc@{}}
		\toprule
		
		\multirow{2}{*}{NB} &
		\multirow{2}{*}{LA} &
		\multirow{2}{*}{LC} &
		\multirow{2}{*}{LL} &
		\multirow{2}{*}{CAR} &
		\multicolumn{7}{c|}{\textbf{MVTec-AD}} &
		\multicolumn{7}{c}{\textbf{VisA}} \\
		
		\cmidrule(lr){6-12}
		\cmidrule(lr){13-19}
		
		& & & & &
		\multicolumn{3}{c|}{\textit{Image-Level}} &
		\multicolumn{4}{c|}{\textit{Pixel-Level}} &
		\multicolumn{3}{c|}{\textit{Image-Level}} &
		\multicolumn{4}{c}{\textit{Pixel-Level}} \\
		
		\midrule
			
			% Baseline
			& & & & &
			\cellcolor{ours!10}98.73 &
			\cellcolor{ours!10}99.29 &
			\cellcolor{ours!10}97.83 &
			\cellcolor{ours!10}97.48 &
			\cellcolor{ours!10}64.82 &
			\cellcolor{ours!10}65.06 &
			\cellcolor{ours!10}93.38 &
			\cellcolor{ours!10}96.23 &
			\cellcolor{ours!10}96.62 &
			\cellcolor{ours!10}92.49 &
			\cellcolor{ours!10}97.98 &
			\cellcolor{ours!10}49.34 &
			\cellcolor{ours!10}53.51 &
			\cellcolor{ours!10}93.41 \\
			
			% NB
			\checkmark & & & & &
			\cellcolor{ours!20}99.09 &
			\cellcolor{ours!20}99.55 &
			\cellcolor{ours!20}98.34 &
			\cellcolor{ours!20}97.74 &
			\cellcolor{ours!20}66.85 &
			\cellcolor{ours!20}66.84 &
			\cellcolor{ours!20}93.81 &
			\cellcolor{ours!30}97.26 &
			\cellcolor{ours!30}97.53 &
			\cellcolor{ours!30}93.83 &
			\cellcolor{ours!10}97.89 &
			\cellcolor{ours!10}50.33 &
			\cellcolor{ours!10}54.45 &
			\cellcolor{ours!10}93.34 \\
			
			% LA
			& \checkmark & & & &
			\cellcolor{ours!15}98.98 &
			\cellcolor{ours!15}99.42 &
			\cellcolor{ours!15}98.15 &
			\cellcolor{ours!15}97.61 &
			\cellcolor{ours!15}65.66 &
			\cellcolor{ours!15}65.55 &
			\cellcolor{ours!15}93.59 &
			\cellcolor{ours!10}96.19 &
			\cellcolor{ours!10}96.53 &
			\cellcolor{ours!10}92.55 &
			\cellcolor{ours!15}98.05 &
			\cellcolor{ours!15}49.84 &
			\cellcolor{ours!15}53.69 &
			\cellcolor{ours!15}93.35 \\
			
			% LC
			& & \checkmark & & &
			\cellcolor{ours!25}99.21 &
			\cellcolor{ours!25}99.54 &
			\cellcolor{ours!25}98.51 &
			\cellcolor{ours!25}97.78 &
			\cellcolor{ours!25}65.24 &
			\cellcolor{ours!25}65.46 &
			\cellcolor{ours!25}93.80 &
			\cellcolor{ours!20}96.88 &
			\cellcolor{ours!20}97.26 &
			\cellcolor{ours!20}93.22 &
			\cellcolor{ours!25}98.20 &
			\cellcolor{ours!25}50.11 &
			\cellcolor{ours!25}54.08 &
			\cellcolor{ours!25}93.66 \\
			
			% LL
			& & & \checkmark & &
			\cellcolor{ours!25}99.19 &
			\cellcolor{ours!25}99.56 &
			\cellcolor{ours!25}98.34 &
			\cellcolor{ours!30}97.83 &
			\cellcolor{ours!30}66.00 &
			\cellcolor{ours!30}66.24 &
			\cellcolor{ours!30}93.66 &
			\cellcolor{ours!25}97.03 &
			\cellcolor{ours!25}97.29 &
			\cellcolor{ours!25}93.40 &
			\cellcolor{ours!35}98.28 &
			\cellcolor{ours!35}50.41 &
			\cellcolor{ours!35}54.25 &
			\cellcolor{ours!35}93.77 \\
			
			% CAR
			& & & & \checkmark &
			\cellcolor{ours!15}98.97 &
			\cellcolor{ours!15}99.40 &
			\cellcolor{ours!15}98.36 &
			\cellcolor{ours!20}97.74 &
			\cellcolor{ours!20}65.99 &
			\cellcolor{ours!20}65.95 &
			\cellcolor{ours!20}93.73 &
			\cellcolor{ours!35}97.42 &
			\cellcolor{ours!35}97.52 &
			\cellcolor{ours!35}94.13 &
			\cellcolor{ours!60}98.54 &
			\cellcolor{ours!60}51.03 &
			\cellcolor{ours!60}54.73 &
			\cellcolor{ours!60}94.87 \\
			
			% NB + LA
			\checkmark & \checkmark & & & &
			\cellcolor{ours!40}99.32 &
			\cellcolor{ours!40}99.67 &
			\cellcolor{ours!40}98.72 &
			\cellcolor{ours!40}97.95 &
			\cellcolor{ours!40}67.40 &
			\cellcolor{ours!40}67.44 &
			\cellcolor{ours!40}94.14 &
			\cellcolor{ours!40}97.52 &
			\cellcolor{ours!40}97.64 &
			\cellcolor{ours!40}94.32 &
			\cellcolor{ours!15}98.01 &
			\cellcolor{ours!15}50.71 &
			\cellcolor{ours!15}54.61 &
			\cellcolor{ours!15}93.76 \\
			
			% NB + LA + LC
			\checkmark & \checkmark & \checkmark & & &
			\cellcolor{ours!70}99.64 &
			\cellcolor{ours!70}99.78 &
			\cellcolor{ours!70}99.23 &
			\cellcolor{ours!70}98.25 &
			\cellcolor{ours!70}67.69 &
			\cellcolor{ours!70}67.97 &
			\cellcolor{ours!70}94.51 &
			\cellcolor{ours!65}98.15 &
			\cellcolor{ours!65}98.35 &
			\cellcolor{ours!65}95.03 &
			\cellcolor{ours!50}98.43 &
			\cellcolor{ours!50}50.98 &
			\cellcolor{ours!50}54.92 &
			\cellcolor{ours!50}94.02 \\
			
			% NB + LA + LC + LL
			\checkmark & \checkmark & \checkmark & \checkmark & &
			\cellcolor{ours!70}\underline{99.64} &
			\cellcolor{ours!70}\underline{99.79} &
			\cellcolor{ours!70}\textbf{99.33} &
			\cellcolor{ours!90}\textbf{98.36} &
			\cellcolor{ours!90}\textbf{69.00} &
			\cellcolor{ours!90}\textbf{68.85} &
			\cellcolor{ours!90}\underline{94.57} &
			\cellcolor{ours!70}\underline{98.21} &
			\cellcolor{ours!70}\underline{98.34} &
			\cellcolor{ours!70}\underline{95.13} &
			\cellcolor{ours!65}\underline{98.61} &
			\cellcolor{ours!65}\underline{51.01} &
			\cellcolor{ours!65}\underline{54.97} &
			\cellcolor{ours!65}\underline{94.31} \\
			
			% Full model
			\checkmark & \checkmark & \checkmark & \checkmark & \checkmark &
			\cellcolor{ours!100}\textbf{99.74} &
			\cellcolor{ours!100}\textbf{99.82} &
			\cellcolor{ours!100}\underline{99.30} &
			\cellcolor{ours!85}\underline{98.34} &
			\cellcolor{ours!85}\underline{68.97} &
			\cellcolor{ours!85}\underline{68.64} &
			\cellcolor{ours!85}\textbf{94.81} &
			\cellcolor{ours!100}\textbf{98.89} &
			\cellcolor{ours!100}\textbf{99.02} &
			\cellcolor{ours!100}\textbf{96.47} &
			\cellcolor{ours!100}\textbf{98.92} &
			\cellcolor{ours!100}\textbf{53.22} &
			\cellcolor{ours!100}\textbf{56.29} &
			\cellcolor{ours!100}\textbf{95.28} \\
			
			\bottomrule
		\end{tabular}
	}
	\label{tab:ablation}
\end{table*}

%  ───────────── Ablation ViTAD─────────────

\begin{table}[t]
  \centering
    \footnotesize
  \setlength{\tabcolsep}{3pt} 
  \caption{Integrating Dinomaly components on ViTAD (\%). Models are trained for 20,000 iterations.}

      \begin{tabular*}{\columnwidth}{@{\extracolsep{\fill}} llccccccc @{}}
    \toprule
  {Method\&Data} & Components & I-ROC & P-ROC & {P-AP} & {P-PRO} \\   \midrule
ViTAD & \textit{reproduced} & 98.27 & 97.49 &58.57 & 91.45 \\ 
 (MVTec-AD)     & $\uparrow$+NB &  98.66 & 97.93  & 61.73 &  92.94 \\
      & ~$\uparrow$+LA & 98.56  & 98.01  & 62.20 & 93.14  \\
      & ~~$\uparrow$+LC &   98.92 & 98.25  &62.16   & 93.05  \\
      & ~~~$\uparrow$+LL &  98.95 & 98.32  & \textbf{62.68} & 93.21  \\
      & ~~~~$\uparrow$+CAR &  \textbf{99.25} & \textbf{98.38}  &  62.18 & \textbf{93.48}  \\ \midrule
ViTAD & \textit{reproduced} & 90.69 & 97.52 & 39.65 & 83.24 \\ 
 (VisA)     & $\uparrow$+all &  \textbf{94.34} & \textbf{98.58}  & \textbf{43.69} &  \textbf{88.69} \\
\bottomrule
    \end{tabular*}%
    
  \label{tab:vitad}%
\vspace{-5pt}
\end{table}%

\begin{table}[t]
	\centering
	\footnotesize
	\setlength{\tabcolsep}{3pt}
	\caption{Controlled comparison under matched settings on MVTec-AD and VisA (\%). All methods use the same frozen DINOv2-R Base backbone, $392^2$ input resolution, and training recipe.}
	\begin{tabular*}{\columnwidth}{@{\extracolsep{\fill}}llcccc@{}}
		\toprule
		Dataset & Method & I-ROC & P-ROC & P-AP & P-PRO \\
		\midrule
		\multirow{5}{*}{\textbf{MVTec-AD}} & RD4AD & 96.62 & 95.94 & 56.38 & 89.47 \\
		& UniAD & 94.66 & 95.98 & 55.60 & 91.31 \\
		& ViTAD & 99.24 & 97.51 & 63.03 & 93.19 \\
		& MambaAD & 98.94 & 97.68 & 67.77 & 94.41 \\
		& \textbf{Dinomaly2} & \textbf{99.80} & \textbf{98.35} & \textbf{69.28} & \textbf{94.76} \\
		\hline
		\multirow{5}{*}{\textbf{VisA}} & RD4AD & 96.03 & 96.39 & 45.31 & 91.96 \\
		& UniAD & 88.13 & 96.09 & 30.24 & 83.61 \\
		& ViTAD & 97.18 & 97.75 & 48.09 & 92.30 \\
		& MambaAD & 94.05 & 97.44 & 42.94 & 91.78 \\
		& \textbf{Dinomaly2} & \textbf{99.15} & \textbf{99.06} & \textbf{52.89} & \textbf{95.54} \\
		\bottomrule
	\end{tabular*}%
	\label{tab:controlled_comparison}%
	\vspace{-5pt}
\end{table}%

\subsubsection{Controlled Comparison under Matched Settings}
Dinomaly2 achieves the best image- and pixel-level results under matched settings on MVTec-AD and VisA (Table~\ref{tab:controlled_comparison}). We reproduce ViTAD, UniAD, RD4AD, and MambaAD using the same frozen DINOv2-R Base backbone, $392^2$ resolution, and training recipe as Dinomaly2, while retaining each method's original reconstruction architecture and objective.

%  ───────────── Ablation Noise ─────────────

\begin{table}[t]
  \centering
  \footnotesize
  \setlength{\tabcolsep}{3pt} 
  \caption{Ablation of Noisy Bottleneck on MVTec-AD (\%). Models are trained for 10,000 iterations with ViT-B, LA, and LC.}
      \begin{tabular*}{\columnwidth}{@{\extracolsep{\fill}} lccccc @{}}

    \toprule
 Noise type & Noise rate & I-ROC & P-ROC & {P-AP} & {P-PRO} \\   \midrule
None & 0 & 99.24 & 97.87 &65.76 & 93.95 \\ \hline
Synthetic & DRAEM & 99.42 & 97.69 & 62.40 & 93.65 \\ \hline
Noisy $\mathcal{B}$ & s=5 & 98.94 & 97.64 &65.23 & 92.88 \\ 
(Feature Jitter) & s=10 & 99.27 & 98.05 &67.85 & 94.03 \\
      & s=20 & 99.58 & 98.13 &68.16 & 94.22 \\
    & s=40 & 99.56 & 98.05 &67.81 & 94.53 \\ \hline

Noisy $\mathcal{B}$ &p=0.1 & 99.56 & 98.25 &67.79 & 94.28 \\
(Dropout) &p=0.2 & 99.64 & 98.25 &67.69 & 94.51 \\
 &p=0.3 & 99.65 & \textbf{98.29} &\textbf{68.28} & 94.49 \\
 &p=0.4 & \textbf{99.72} & \textbf{98.29} &67.76 & 94.47 \\
 &p=0.5 & 99.63 & 98.25 &67.54 & \textbf{95.01} \\
 \hline
Noisy $\mathcal{D}$ &p=0.2 & 99.02 & 97.71 &66.00 & 93.75 \\
Noisy $\mathcal{B}+\mathcal{D}$ &p=0.2 & 99.26 & 98.06 &67.82 & 94.25 \\

\bottomrule
    \end{tabular*}%

  \label{tab:dropout}%
\end{table}%

%  ───────────── Ablation LA ─────────────

\begin{table}[t]
  \centering
  \footnotesize
  \setlength{\tabcolsep}{3pt} 
  \caption{Ablation of Unfocused Attention on MVTec-AD (\%). Models are trained for 10,000 iterations with ViT-B, NB, and LC.}
  \begin{tabular*}{\columnwidth}{@{\extracolsep{\fill}} lccccc @{}}

    \toprule
Spatial Mixer & I-ROC & P-ROC & {P-AP} & {P-PRO} \\   \midrule
ConvBlock 1x1 & 99.45 & 98.05 &65.35 & 94.01 \\
ConvBlock 3x3 & 99.43 & 98.02 &65.85 & 94.14 \\
ConvBlock 5x5 & 99.42 & 98.07 &67.25 & 94.29 \\
Softmax Attention & 99.49 & 98.17 &67.65 & 94.33 \\ 
% Attention w/neighbor-mask n=1 & 99.56 & 98.25 &67.79 & 94.28 \\
% Attention w/neighbor-mask n=3 & 99.59 & 98.17 &67.69 & 94.38 \\ \hline
Unfocused Linear Attention & \textbf{99.64} & \textbf{98.25} &\textbf{67.97} & \textbf{94.51} \\
\bottomrule
    \end{tabular*}%
  \label{tab:attention}%
\end{table}%

%  ───────────── Ablation LC ─────────────

\begin{table}[t]
  \centering
  \footnotesize
  \setlength{\tabcolsep}{3pt} 
  \caption{Ablation of Loose Constraint on MVTec-AD (\%). Models are trained for 10,000 iterations with ViT-B, NB, and LA.}
  \begin{tabular*}{\columnwidth}{@{\extracolsep{\fill}} lccccc @{}}
    \toprule
Constraints & I-ROC & P-ROC & {P-AP} & {P-PRO} \\   \midrule
layer2layer (last 1) & 98.80 & 97.15 &60.16 & 92.67 \\ 
layer2layer (dense, every 1) & 99.32 & 97.95 &67.40 & 94.14 \\
layer2layer (sparse, every 2) & 99.47  & 98.01  & 67.15  & 94.32  \\
layer2layer (sparse, every 4) & 99.43 & 97.94 &66.03 & 94.10 \\ \hline
group2group (1 group) & 99.61 & 98.19 &65.22 & 94.15 \\
group2group (2 group) & \textbf{99.64} & \textbf{98.25} &\textbf{67.69} & \textbf{94.51} \\
group2group (4 group) & 99.47 & 98.11 &67.64 & 94.32 \\
\bottomrule
    \end{tabular*}%

  \label{tab:group}%
\vspace{-5pt}
\end{table}%

\begin{table}[t]
	\centering
	\footnotesize
	\setlength{\tabcolsep}{3pt}
	\caption{Ablation of target transformations (\%). All variants retain NB, LA, LC, and LL and are trained for 10,000 iterations. Image CLS concatenation is applied to both encoder and decoder features. Category-average CLS subtraction requires class labels for both training and test images.}
	\begin{tabular*}{\columnwidth}{@{\extracolsep{\fill}}ll>{\tiny}lcccc@{}}
		\toprule
		& \makecell{Target\\trans.} & \footnotesize{Variant} & I-ROC & P-ROC & P-AP & P-PRO \\
		\midrule
		\multirow{9}{*}{\rotatebox{90}{\textbf{MVTec-AD}}}
		& None & None & 99.64 & 98.36 & 69.00 & 94.57 \\
		\hhline{~------}
		& \multirow{3}{*}{Norm.} & LayerNorm & 99.64 & 98.37 & 68.77 & 94.60 \\
		& & InstanceNorm & 99.41 & 97.73 & 66.52 & 93.26 \\
		& & RMSNorm & 99.63 & 98.36 & 69.07 & 94.61 \\
		\hhline{~------}
		& \multirow{2}{*}{\makecell{Non-\\CLS}} & Patch mean sub. & 99.03 & 97.82 & 65.00 & 92.84 \\
		& & Learnable center sub. & 99.33 & 97.92 & 65.62 & 94.29 \\
		\hhline{~------}
		& \multirow{3}{*}{\makecell{CLS-\\based}} & Image CLS sub. & \textbf{99.74} & 98.34 & 68.97 & 94.81 \\
		& & Image CLS concat. & 99.70 & \textbf{98.44} & 68.86 & 94.73 \\
		& & Category CLS sub. & 99.72 & 98.32 & \textbf{68.99} & \textbf{95.27} \\
		
		\midrule
		\multirow{9}{*}{\rotatebox{90}{\textbf{VisA}}}
		& None & None & 98.21 & 98.61 & 51.01 & 94.31 \\
		\hhline{~------}
		& \multirow{3}{*}{Norm.} & LayerNorm & 98.42 & 98.70 & 50.49 & 94.30 \\
		& & InstanceNorm & 96.21 & 97.36 & 48.60 & 88.52 \\
		& & RMSNorm & 96.85 & 98.56 & 48.22 & 94.04 \\
		\hhline{~------}
		& \multirow{2}{*}{\makecell{Non-\\CLS}} & Patch mean sub. & 94.41 & 96.69 & 45.72 & 87.10 \\
		& & Learnable center sub. & 98.49 & 98.85 & 50.53 & 94.96 \\
		\hhline{~------}
		& \multirow{3}{*}{\makecell{CLS-\\based}} & Image CLS sub. & 98.89 & 98.92 & \textbf{53.22} & 95.28 \\
		& & Image CLS concat. & 98.76 & \textbf{98.94} & 51.68 & 95.06 \\
		& & Category CLS sub. & \textbf{98.99} & 98.86 & 52.45 & \textbf{95.46} \\
		\bottomrule
	\end{tabular*}%
	\label{tab:recenter_variants}%
	\vspace{-5pt}
\end{table}%

\subsubsection{Ablation on Other Architectures} To demonstrate the generalizability of our proposed components beyond the Dinomaly framework, we systematically integrate each module into ViTAD~\citep{zhang2023exploring}, a representative Transformer-based UAD method. As shown in Table~\ref{tab:vitad}, each Dinomaly2 component provides consistent improvements. The progressive integration yields substantial gains: starting from 98.27\% to 99.25\% I-AUROC—a total improvement of 0.98\% I-AUROC and 2.03\% AUPRO. This transferability analysis validates that our design principles are not framework-specific optimizations but applicable to diverse UAD architectures.

\subsubsection{Individual Component Analysis} 

\textbf{Noisy Bottleneck.} We conduct ablations for Noisy Bottleneck as presented in Table~\ref{tab:dropout}. Dinomaly2 performs well across a range of dropout rates, with the best results obtained between 0.2 and 0.4. We also evaluate the effectiveness of other types of noise, such as synthetic image anomaly (DRAEM-like~\citep{zavrtanik2021draem}), feature jitter~\citep{you2022unified} (Gaussian noise with \textit{scale} to control the noise magnitude). Both Dropout and feature jitter provide effective noise injection, but Dropout is less sensitive to its hyperparameter and requires no additional modules.  Furthermore, applying dropout in the decoder (Noisy $\mathcal{D}$: 99.02\%) or both locations (Noisy $\mathcal{B}+\mathcal{D}$: 99.26\%) harms the performance, validating our theoretical interpretation: the Noisy Bottleneck functions as an information constraint rather than a generic regularization technique.

\textbf{Unfocused Attention.} As shown in Table~\ref{tab:attention}, Unfocused Linear Attention consistently outperforms alternatives including standard convolutions and Softmax Attention. Linear Attention surpasses vanilla Softmax Attention using less computation, validating our counterintuitive approach of leveraging attention's inability to focus as a desirable property.

\textbf{Loose Reconstruction.} We quantitatively examine different reconstruction schemes presented in Fig.~\ref{fig5}. Table~\ref{tab:group} shows that group-to-group Loose Constraint outperforms layer-to-layer supervision. Our 2-group Loose Constraint strategy (grouping into shallow and deep semantic levels) achieves optimal performance (99.64\% I-AUROC), outperforming both single-group and more granular 4-group strategies. This supports our hypothesis that appropriate semantic grouping provides the decoder with sufficient degrees of freedom without losing important hierarchical information.

\textbf{Context-Aware Recentering.} The target-transformation ablations support contextual information, rather than generic normalization or centering, as the main source of CAR's benefit (Table~\ref{tab:recenter_variants}). All variants retain NB, LA, LC, and LL; reconstruction targets remain detached except for the shared center in the learnable-center control. The differences are clearest on VisA, as MVTec-AD is already close to saturation under this configuration.

Category-average CLS performs comparably to image-specific CLS but requires class labels for both training and test images. Image CLS concatenation is also an effective alternative, but doubles the feature dimension. We therefore use image-specific CLS subtraction as the parameter-free and label-free default. The gains from CLS conditioning provide empirical support for the hypothesis of cross-class contextual ambiguity.

\begin{figure}[!t]
	\centerline{\includegraphics[width=0.95\linewidth]{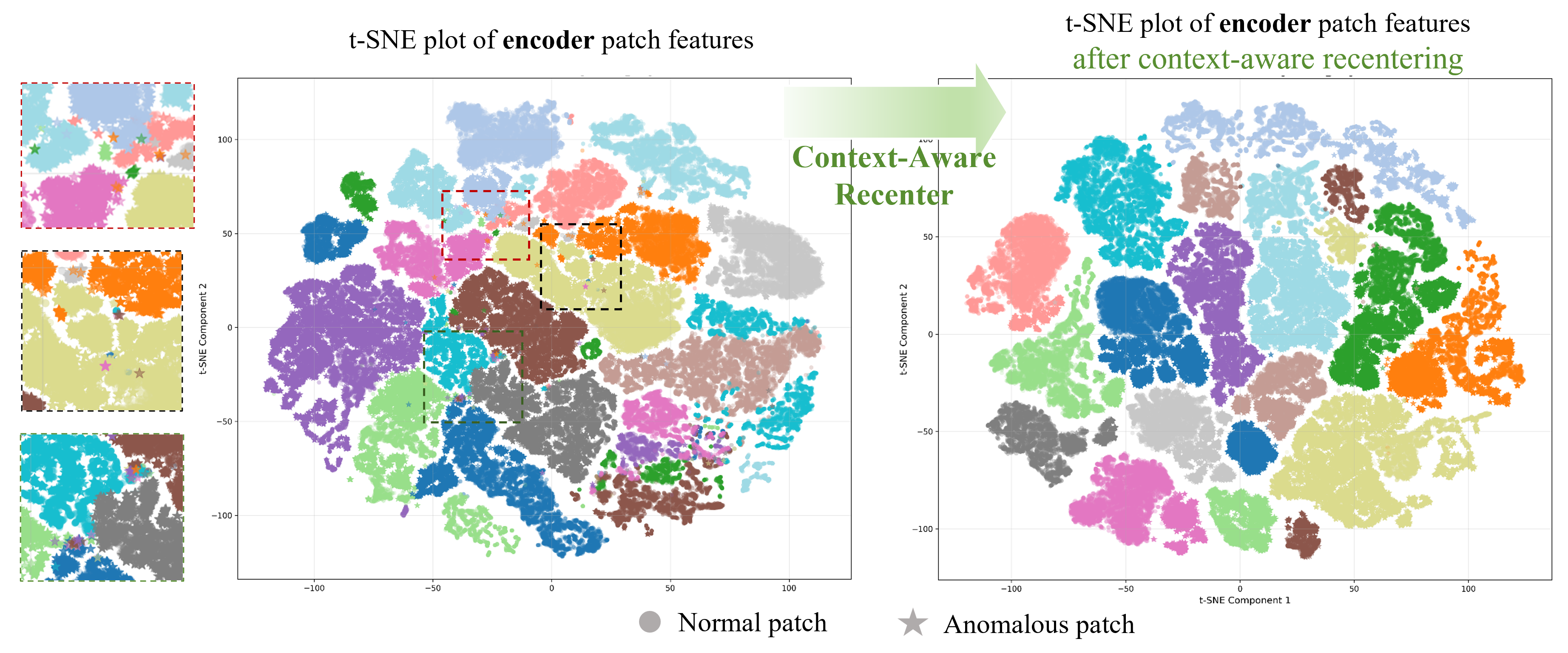}}
	\setlength{\abovecaptionskip}{2pt} 
	\caption{t-SNE visualization of encoder patch features on MVTec-AD. Different colors represent different classes. }
	\label{fig:tsne}
	\vspace{-10pt}
\end{figure}

\begin{table*}[!t]
  \centering
  \scriptsize
  \caption{Model size scaling analysis across model sizes on MVTecAD, VisA, and Real-IAD (\%). FPS (throughput, images per second) is measured on NVIDIA GeForce RTX {3090/4090} with batch size=16. Training time are measured on an NVIDIA GeForce RTX 3090 and reported per 10,000 iterations.}
   \resizebox{0.99\linewidth}{!}{
    \begin{tabular}{lccccccccccccccc}
    \toprule
\multirow{2}[2]{*}{Method} & \multirow{2}[2]{*}{Params} & \multirow{2}[2]{*}{MACs} & \multirow{2}[2]{*}{\makecell[c]{FPS\\3090/4090}} & \multirow{2}[2]{*}{\makecell[c]{Train \\ Mem.}} & \multirow{2}[2]{*}{\makecell[c]{Train Time\\(10k iters)}} & \multicolumn{3}{c}{\textbf{MVTec-AD}} & \multicolumn{3}{c}{\textbf{VisA}} & \multicolumn{4}{c}{\textbf{Real-IAD}} \\
\cmidrule(r){7-9} \cmidrule(l){10-12}  \cmidrule(l){13-16}
& & & & & & \multicolumn{1}{c}{I-ROC} & \multicolumn{1}{c}{P-ROC} & \multicolumn{1}{c}{P-PRO} & \multicolumn{1}{c}{I-ROC} & \multicolumn{1}{c}{P-ROC} & \multicolumn{1}{c}{P-PRO} &  \multicolumn{1}{c}{O-ROC} &  \multicolumn{1}{c}{I-ROC} & \multicolumn{1}{c}{P-ROC} & \multicolumn{1}{c}{P-PRO} \\ \midrule
\rowcolor{ours!30} Dinomaly2-S & 37.3M & 26.2G & 153.6/253.0 & 5.3GB & 0.59h & 99.62 & 98.35 & 94.91 & 98.69 & 98.96  &94.98 & 93.71  & 90.25 & 98.99 & 95.32\\
\rowcolor{ours!60} Dinomaly2-B & 146.6M & 103.5G & 58.1/103.8 & 10.6GB & 1.41h & 99.80& 98.35 &94.76 & 99.15 & 99.06 & 95.54 & 94.30 & 91.57 & 99.16 & 95.99 \\
\rowcolor{ours!90} Dinomaly2-L & 410.7M & 272.9G & 24.2/45.2 & 15.0GB & 2.80h &  99.89 & 98.55 & 95.08 & 99.29 & 99.15 & 95.38 & 94.93 & 92.07 & 99.24 & 96.24 \\
\bottomrule
\end{tabular}
}
\label{tab:arch}
\end{table*}

\begin{table}[t]
  \centering
  \tiny
  \caption{Input resolution scaling comparison on MVTec-AD (\%).  \dag: default of the method. Note that previous methods often yield degradation with larger input sizes.}
  \begin{tabular*}{\columnwidth}{@{\extracolsep{\fill}} lccccccc @{}}
    \toprule
 Method & Input Size & I-ROC & P-ROC & \textbf{P-AP} & \textbf{P-PRO} \\   \hline
RD4AD &256$\times$256\dag  & \textbf{94.6} & \textbf{96.1} &48.6 & \textbf{91.1} \\
&320$\times$320 & 93.2 & 95.7 &\textbf{55.1} & \textbf{91.1} \\
&384$\times$384 & 91.9 & 94.9 &52.1 & 90.8 \\
 \hline
 ReContrast &256$\times$256\dag  & \textbf{98.3} & \textbf{97.1} &60.2 & 93.2 \\
&320$\times$320 & 98.2 & 96.8 &\textbf{61.8} & \textbf{93.3} \\
&384$\times$384 & 95.2 & 96.5 &57.7 & 92.6 \\
 \hline
 ViTAD &256$\times$256\dag  & \textbf{98.3} & \textbf{97.7} &55.3 & 91.4 \\
&320$\times$320 & \textbf{98.3} & 97.6 & 61.3 & \textbf{92.4} \\
&384$\times$384 & 97.8 & 97.5 &\textbf{62.5} & \textbf{92.4} \\
 \hline
\textbf{Dinomaly2-B} & 224$\times$224 & 99.7 & 98.3 & 65.2 & 93.3 \\
&252$\times$252 & \textbf{99.8} & 98.2 &65.3 & 93.8 \\
&280$\times$280 & \textbf{99.8} & 98.3 &66.7 & 94.2 \\
& 336$\times$336 & \textbf{99.8} & 98.3 &68.0 & 94.4 \\
& 392$\times$392\dag & \textbf{99.8} & \textbf{98.4} &69.3 & 94.8 \\
& 448$\times$448 & \textbf{99.8} & \textbf{98.4} &\textbf{70.3} & \textbf{95.1} \\
& 518$\times$518 & \textbf{99.8} & 98.3 & \textbf{70.3} & 94.9 \\
\bottomrule
    \end{tabular*}%
  \label{tab:isize}%
\end{table}%

The t-SNE visualization of encoder patch features on MVTec-AD further illustrates how CAR reduces cross-class overlap (Fig.~\ref{fig:tsne})~\citep{maaten2008visualizing}. Before recentering, anomalous patches from one class often overlap with normal patches from another. Subtracting each image's own CLS token produces more distinct class-wise feature clusters while preserving within-class separation between normal and anomalous patches.

\begin{figure*}[t]
\centerline{\includegraphics[width=\textwidth]{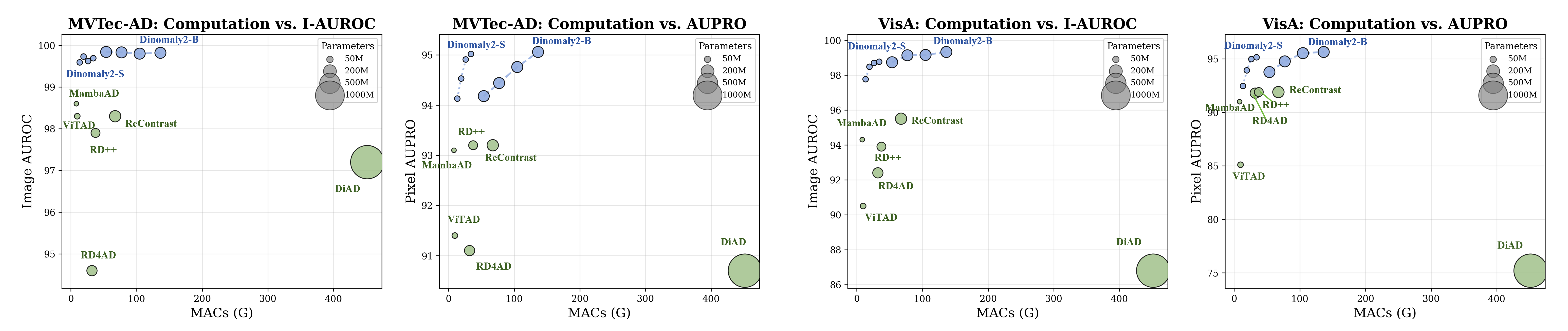}}
\setlength{\abovecaptionskip}{2pt} 
\caption{Computation--performance trade-offs on MVTec-AD and VisA. The horizontal axis denotes MACs, bubble size represents parameter count, and connected points show Dinomaly2-S and Dinomaly2-B at input resolutions $280^2$, $336^2$, $392^2$, and $448^2$.}
\label{fig:scaling}
%\vspace{-5pt}
\end{figure*}

\subsection{Scaling Properties of Dinomaly2}

\subsubsection{Model Size Scaling} 
Unlike previous UAD methods that often reported diminishing or negative returns from larger models~\citep{zhang2023exploring,he2024mambaad}, Dinomaly2 demonstrates clear positive scaling behavior across model sizes. As shown in Table~\ref{tab:arch}, scaling from ViT-S (37.3M parameters, including bottleneck and decoder) to ViT-B (146.6M) and further to ViT-L (410.7M) improves overall performance.

Importantly, this scaling is achieved with reasonable and flexible computational cost. Dinomaly2-S already produced state-of-the-art results while maintaining a high inference speed of 253 images/second on a consumer-level GPU (NVIDIA RTX4090). As reported in Table~\ref{tab:arch}, Dinomaly2-S/B/L require 5.3/10.6/15.0 GB of peak training memory and 0.59/1.41/2.80 hours per 10,000 training iterations, respectively, measured on an NVIDIA RTX 3090. While Dinomaly2-L requires higher computational resources, its throughput of 45 images/second remains practical for many real-world applications. This scalability enables practitioners to choose appropriate model sizes based on their performance-efficiency trade-offs.

\begin{table*}[t]
\footnotesize
  \centering
  \caption{Comprehensive evaluation of pre-trained Vision Transformer foundations on MVTec-AD (\%). The patch size of DINOv2, DINOv2-R, and TIPS is 14x14; others are 16x16. MIM: masked image modeling. CL: contrastive learning. VLM: vision language modeling. Dv1 and Dv2 denote Dinomaly and Dinomaly2. }
    \begin{tabular}{llccccccc}
    \toprule
    \makecell[l]{Backbone} & \makecell[l]{Pre-train\\Type} & \makecell[c]{{ImageNet}\\{Linear-prob}} & {Method} & \makecell[c]{Image\\Size} & I-ROC & P-ROC & P-AP & P-PRO \\
    \midrule
MAE~\citep{he2022masked} & MIM& 68.0 & ViTAD & $256^2$ & 95.3 & 97.4 & 53.0 & 90.6 \\
DINO~\citep{caron2021emerging} & CL& 78.2 & ViTAD & $256^2$ & 98.3 & 97.7 & 55.3 & 91.4 \\
DINOv2~\citep{oquab2023dinov2} & CL+MIM& 84.5 & ViTAD & $224^2$ & 98.7 & 97.6 & 55.3 & 92.0 \\
DINOv2-R~\citep{darcet2023vision} & CL+MIM& 84.8 & ViTAD & $224^2$ & 98.5 & 97.4 & 54.5 & 92.1 \\
    \midrule
DeiT~\citep{touvron2021training} & Supervised& -& Dv1 & $224^2$ & 97.65 & 97.80 & 62.58 & 89.98 \\
MAE~\citep{he2022masked} & MIM& 68.0 & Dv1 & $224^2$ & 97.25 & 97.78 & 63.00 & 90.95 \\
BEiTv2~\citep{peng2002beitv2} & MIM& 80.1 & Dv1 & $224^2$ & 97.70 & 97.61 & 59.79 & 90.10 \\
D-iGPT~\citep{ren2023rejuvenating} & MIM& 80.5 & Dv1 & $224^2$ & 99.21 & 98.08 & 60.05 & 91.78 \\
MOCOv3~\citep{chen2021empirical} & CL& 76.7 & Dv1 & $224^2$ & 98.74 & 98.05 & 63.36 & 91.13 \\
DINO~\citep{caron2021emerging} & CL& 78.2 & Dv1 & $224^2$ & 99.20 & 98.16 & 64.16 & 92.02 \\
iBOT~\citep{zhou2021ibot} & CL+MIM& 79.5 & Dv1 & $224^2$ & 99.31 & 98.25 & 64.01 & 91.68 \\
DINOv2~\citep{oquab2023dinov2} & CL+MIM&84.5 & Dv1 & $224^2$ & 99.26 & 97.95 & 62.27 & 92.80 \\
DINOv2-R~\citep{darcet2023vision} & CL+MIM& 84.8& Dv1 & $224^2$ & 99.34 & 98.09 & 63.04 & 92.59 \\ 
\midrule
DeiT~\citep{touvron2021training} & Supervised& - & Dv1& $448^2$ & 98.19 & 97.93 & 68.98 & 91.45 \\
DeiT~\citep{touvron2021training} & Supervised& - & Dv2& $448^2$ & 98.41 & 97.99 & 68.31 & 91.85 \\
D-iGPT~\citep{ren2023rejuvenating} & MIM& 80.5 & Dv1 & $448^2$ & 98.75 & 98.30 & 65.77 & 92.34 \\
MOCOv3~\citep{chen2021empirical} & CL& 76.7 & Dv1 & $448^2$ & 98.47 & 98.52 & 70.99 & 92.83 \\
DINO~\citep{caron2021emerging} & CL& 78.2& Dv1 & $448^2$ & 98.97 & 98.52 & 70.89 & 93.48 \\
TIPS~\citep{maninis2024tips} & VLM& 81.8 & Dv1 & $392^2$ & 99.53 & 98.39 & 67.17 & 94.62 \\
TIPS~\citep{maninis2024tips} & VLM& 81.8 & Dv2 & $392^2$ & 99.69 & 98.18 & 64.63 & 93.81 \\

iBOT~\citep{zhou2021ibot} & CL+MIM&79.5 & Dv1 & $448^2$ & 99.22 & 98.60 & 70.78 & 93.33 \\
iBOT~\citep{zhou2021ibot} & CL+MIM& 79.5 & Dv2 & $448^2$ & 99.38 & 98.27 & 68.76 & 91.73 \\
DINOv2~\citep{oquab2023dinov2} & CL+MIM& 84.5 & Dv1 & $392^2$ & 99.55 & 98.26 & 68.35 & 94.83 \\
DINOv2~\citep{oquab2023dinov2} & CL+MIM& 84.5 & Dv2 & $392^2$ & 99.83 & 98.46 & 69.83 & 95.16 \\
DINOv2-R~\citep{darcet2023vision} & CL+MIM& 84.8 & Dv1 & $392^2$ & 99.60 & 98.35 & 69.29 & 94.79 \\
\textbf{DINOv2-R} ~\citep{darcet2023vision} & CL+MIM& 84.8 & Dv2 & $392^2$ & 99.80 & 98.35 & 69.28 & 94.76 \\

DINOv3$_{\text{ViT-S}}$~\citep{simeoni2025dinov3} & CL+MIM& - & Dv2 & $448^2$ & 99.50 & 98.00 & 70.78 & 94.18 \\
DINOv3$_{\text{ViT-B}}$~\citep{simeoni2025dinov3} & CL+MIM& - & Dv2 & $448^2$ & 99.82 & 98.40 & 71.88 & 94.90 \\
DINOv3$_{\text{ViT-L}}$~\citep{simeoni2025dinov3} & CL+MIM& - & Dv2 & $448^2$ & 99.85 & 98.86 & 73.04 & 95.58 \\
\bottomrule
\end{tabular}

\label{tab:pretrain}
\end{table*}

\subsubsection{Input Resolution Scaling}
Table~\ref{tab:isize} reveals a sharp contrast between Dinomaly2 and existing methods regarding input size scaling. Previous methods such as RD4AD and ReContrast suffer performance degradation when input resolution increases (e.g., RD4AD drops from 94.6\% to 91.9\% I-AUROC when scaling from $256^2$ to $384^2$).
Dinomaly2 already achieves SOTA image-level detection at $224^2$ (99.7\%), demonstrating its effectiveness at lower input resolutions. More importantly, Dinomaly2 benefits from larger input sizes: pixel-level performance shows substantial gains (from 65.2\% AP to 70.3\% AP), demonstrating that our method effectively leverages higher resolution for more precise anomaly localization. 
Although the DINOv2 backbone is pre-trained at $224\times224$ followed by a high-resolution training stage at $518\times518$, these results indicate that Dinomaly2 operates effectively across multiple resolutions through positional-embedding interpolation.

\subsubsection{Performance under Comparable Computational Budgets}
Dinomaly2 achieves favorable performance--computation trade-offs across compact configurations on MVTec-AD and VisA (Fig.~\ref{fig:scaling}). Dinomaly2-S with 37.3M parameters and 13.4G MACs consistently outperforms the similarly sized ViTAD. Dinomaly2-B with 146.6M parameters and 53.7G MACs also outperforms ReContrast while requiring fewer parameters and MACs.

\begin{table*}[t]
  \centering
  \footnotesize
  \setlength{\tabcolsep}{3pt}
  \caption{Controlled comparison across frozen backbones (\%). Raw Dinomaly denotes the raw Dinomaly baseline without any proposed components (the first row of Table~\ref{tab:ablation}). All models are trained for 10,000 iterations using the same input resolution and training recipe.}
\begin{tabular*}{\textwidth}{@{\extracolsep{\fill}}llcccccccc@{}}
\toprule
Backbone & Method & \multicolumn{4}{c}{\textbf{MVTec-AD}} & \multicolumn{4}{c}{\textbf{VisA}} \\
\cmidrule(lr){3-6} \cmidrule(l){7-10}
& & I-ROC & P-ROC & P-AP & P-PRO & I-ROC & P-ROC & P-AP & P-PRO \\
\midrule
DeiT~\citep{touvron2021training} & ViTAD & 97.06 & 97.10 & 64.13 & 88.71 & 90.37 & 97.31 & 40.94 & 84.43 \\
& Raw Dinomaly & 96.07 & 96.93 & 64.36 & 88.53 & 89.94 & 97.47 & 42.80 & 85.76 \\
& \textbf{Dinomaly2} & \textbf{98.29} & \textbf{98.07} & \textbf{69.11} & \textbf{91.89} & \textbf{95.58} & \textbf{98.17} & \textbf{47.47} & \textbf{91.17} \\
\midrule
iBOT~\citep{zhou2021ibot} & ViTAD & 98.02 & 98.02 & 68.39 & 90.77 & 93.12 & 98.22 & 44.21 & 89.74 \\
& Raw Dinomaly & 97.99 & 98.18 & {68.64} & 91.56 & 93.56 & 98.23 & 47.95 & 90.55 \\
& \textbf{Dinomaly2} & \textbf{99.21} & \textbf{98.35} & \textbf{69.05} & \textbf{92.18} & \textbf{96.05} & \textbf{98.73} & \textbf{50.91} & \textbf{92.70} \\
\midrule
DINOv2-R~\citep{darcet2023vision} & ViTAD & 99.22 & 97.48 & 62.82 & 93.13 & 97.11 & 97.61 & 48.00 & 92.15 \\
& Raw Dinomaly & 98.73 & 97.48 & 64.82 & 93.38 & 96.23 & 97.98 & 49.34 & 93.41 \\
& \textbf{Dinomaly2} & \textbf{99.74} & \textbf{98.34} & \textbf{68.97} & \textbf{94.81} & \textbf{98.89} & \textbf{98.92} & \textbf{53.22} & \textbf{95.28} \\
\midrule
DINOv3~\citep{simeoni2025dinov3} & ViTAD & 99.04 & 97.56 & 67.16 & 93.42 & 97.30 & 98.35 & 52.54 & 93.54 \\
& Raw Dinomaly & 98.48 & 97.37 & 68.00 & 93.35 & 96.61 & 98.17 & 51.72 & 93.72 \\
& \textbf{Dinomaly2} & \textbf{99.66} & \textbf{98.23} & \textbf{71.74} & \textbf{94.71} & \textbf{98.82} & \textbf{98.78} & \textbf{53.43} & \textbf{95.22} \\
\bottomrule
\end{tabular*}
  \label{tab:backbone_controlled}
%  \vspace{-5pt}
\end{table*}

\subsubsection{Pre-trained Foundation Models}
The representation quality of frozen backbones plays a pivotal role in unsupervised anomaly detection performance. To systematically investigate this relationship, we conduct extensive experiments across diverse pre-training paradigms. DeiT ~\citep{touvron2021training} employs supervised learning on ImageNet through knowledge distillation from CNNs. The masked image modeling (MIM) family includes MAE~\citep{he2022masked} (pixel-level reconstruction), BEiTv2~\citep{peng2002beitv2} (VQ-GAN and CLIP token prediction), and D-iGPT~\citep{ren2023rejuvenating} (CLIP feature prediction). Contrastive learning (CL) approaches are represented by MOCOv3~\citep{chen2021empirical} and DINO~\citep{caron2021emerging}, which optimize feature similarity for augmented views of the same image. The most advanced models combine both strategies: iBot~\citep{zhou2021ibot}, DINOv2~\citep{oquab2023dinov2}, and its register-enhanced variant DINOv2-R ~\citep{darcet2023vision}. Additionally, we evaluate two recently released foundation models: TIPS~\citep{maninis2024tips}, which incorporates spatial-aware text-image pre-training, and DINOv3~\citep{simeoni2025dinov3}, the latest successor to DINOv2.

An important consideration emerges regarding input resolution generalization. Most models are pre-trained with the image resolution of $224^2$, except that DINOv2, DINOv2-R, TIPS, and DINOv3 have an additional high-resolution training phase with $518^2$. Applying a frozen encoder at a resolution different from its pre-training resolution may affect feature quality. For comparison, we use a $28\times28$ patch grid by default, corresponding to input resolutions of $392^2$ for ViT-B/14 and $448^2$ for ViT-B/16. We also evaluate our model with $224^2$ inputs. 

The results are presented in Table~\ref{tab:pretrain}. Within Dinomaly or Dinomaly2, nearly all foundation models yield SOTA-level results with I-AUROC higher than 98\%. Employing MAE produces the worst results, which aligns with known limitations of masked auto-encoders for downstream tasks without fine-tuning. This finding corroborates previous observations in ViTAD~\citep{zhang2023exploring}, where MAE as backbone yielded suboptimal results (95.3\%), attributed to weak semantic expression from its pixel-level reconstruction objective.
Models combining CL and MIM consistently outperform single-strategy counterparts, demonstrating the synergistic benefits of multi-objective pre-training. Higher input resolution benefits pixel-level performance for most backbones, but can reduce image-level performance for models pre-trained at $224^2$. D-iGPT, DINO, and iBOT produce results comparable to DINOv2 at $224^2$. This observation suggests that higher-resolution pre-training may also benefit these models.

To distinguish reconstruction-design gains from backbone effects, we compare ViTAD, the raw Dinomaly baseline without any proposed components, and Dinomaly2 across four frozen backbones. All models are trained for 10,000 iterations at the same $392^2$ input resolution. As shown in Table~\ref{tab:backbone_controlled}, Dinomaly2 with DeiT, an early ViT backbone pre-trained with ImageNet supervision, approaches previously reported state-of-the-art image-level performance while substantially outperforming ViTAD and the raw Dinomaly baseline using the same backbone. The results with iBOT, DINOv2-R, and DINOv3 further demonstrate the effectiveness of the proposed design across different representations.

Remarkably, we observe a strong correlation ($\mathcal{R}^2$=0.91) between anomaly detection performance and ImageNet linear-probing accuracy of the foundation model. This suggests that better visual representations translate directly to improved anomaly detection capability, providing a simple heuristic for backbone selection.
The recent release of DINOv3~\citep{simeoni2025dinov3} provides a compelling real-world validation of this predictive principle. As the most advanced foundation model for dense prediction, DINOv3 achieves exceptional pixel-level localization performance in our framework: 71.9/94.9\% in P-AP/AUPRO with ViT-B and 73.0/95.6\% with ViT-L.

 \section{Conclusion} 
 We presented Dinomaly2, a minimalist unified framework for unsupervised anomaly detection. Its five key elements collectively achieve strong performance across diverse data modalities, task settings, and application domains with a shared core architecture and lightweight adaptations. Extensive experiments on various benchmarks demonstrate its advantages over specialized methods. The combination of architectural simplicity and broad applicability positions Dinomaly2 as a practical framework for a wide range of real-world anomaly detection settings.

\section*{Declarations}

\textbf{Ethical Approval.} Not applicable. This article does not contain any studies with human participants or animals.

\noindent \textbf{Consent to Participate.} Not applicable.

\noindent \textbf{Consent for Publication.} Not applicable.

\noindent \textbf{Conflict of Interest.} The authors declare no competing interests.

\noindent \textbf{Funding.} The authors acknowledge supports from National Key Research and Development Program of China (2022YFC2405200, 2025YFC2426300), National Natural Science Foundation of China (U22A2051, 82027807, 82572314, 62271246, 82572313, 62506036), Tsinghua-Foshan Innovation Special Fund (2021THFS0104), the Institute for Intelligent Healthcare, Tsinghua University (2022ZLB001), the Science and Technology Commission of Shanghai Municipality (Nos.24511104100), and the Institute of Digital Medicine, City University of Hong Kong (Project 9229503).

\noindent \textbf{Data Availability.} The datasets used in this study are publicly available, including \href{https://www.mvtec.com/company/research/datasets/mvtec-ad}{MVTec-AD}, \href{https://www.mvtec.com/research-teaching/datasets/mvtec-loco-ad}{MVTec LOCO AD}, \href{https://github.com/amazon-science/spot-diff}{VisA}, \href{https://www.kaggle.com/datasets/thtuan/btad-beantech-anomaly-detection}{BTAD}, \href{https://github.com/stepanje/MPDD}{MPDD}, \href{https://realiad4ad.github.io/Real-IAD/}{Real-IAD}, \href{https://grainnet.github.io/MANTA}{MANTA}, \href{https://www.mvtec.com/company/research/datasets/mvtec-3d-ad}{MVTec3D}, \href{https://github.com/ZZZBBBZZZ/MulSen-AD}{MulSen-AD}, \href{https://drive.google.com/file/d/1Q33X6UMS_2rfdOlHq-Levf7Df7z3tUKp/view?usp=sharing}{Uni-Medical}, \href{https://miad-2022.github.io/}{MIAD}, \href{https://www.kaggle.com/datasets/dayaalex/drone-anomaly}{Drone-Anomaly}.  Code will be available at \href{https://github.com/guojiajeremy/Dinomaly}{URL}.

\noindent \textbf{Author Contributions.} Conceptualization: J. Guo, S. Lu, L. Fan, Z. Li; Method and implementation: J. Guo; Resources: L. Fan, Z. Li, D. Di, Y. Song, W. Zhu, H. Yan, H. Liao;  Draft preparation: J. Guo; Review and editing: L. Fan, Y. Song, W. Zhang, H. Yan, F. Chen, H. Li, H. Liao; Supervision: F. Chen, H. Li, H. Liao. All authors have approved the manuscript.


\bibliography{sn-bibliography}% common bib file

\end{document}